\documentclass[10pt,twocolumn,letterpaper]{article}

\usepackage[accsupp]{axessibility}
\usepackage{cvpr}              

\usepackage{graphicx}
\usepackage{amsmath}
\usepackage{amssymb}
\usepackage{booktabs}
\usepackage{wrapfig}
\usepackage{float}
\usepackage{multirow}
\usepackage{amsfonts}
\usepackage{bbm}
\usepackage{accents}
\usepackage{derivative}
\usepackage[pagebackref,breaklinks,colorlinks]{hyperref}

\usepackage[capitalize]{cleveref}
\crefname{section}{Sec.}{Secs.}
\Crefname{section}{Section}{Sections}
\Crefname{table}{Table}{Tables}
\crefname{table}{Tab.}{Tabs.}

\def\cvprPaperID{6} 
\def\confName{CVPR}
\def\confYear{2022}

\begin{document}

\title{Multi-level Domain Adaptation for Lane Detection}

\author{
   {Chenguang Li}\thanks{Equal contribution.}${~~^{1}}$, 
   {Boheng Zhang}\footnotemark[1]
   \thanks{This work was done during internship at SenseTime Research.}${~~^{1,2}}$,
   {Jia Shi}\footnotemark[2]${~~^{1,3}}$,
   {Guangliang Cheng}\thanks{Guangliang Cheng is the corresponding author.}${^{~~1,4}}$\\
   ${^1}${SenseTime Research} \quad
   ${^2}${Tsinghua University} \quad
   ${^3}${Robotics Institute, Carnegie Mellon University} \\
   ${^4}${Shanghai AI Laboratory} \\
   {\tt\small{lichenguang@senseauto.com, zbh17@mails.tsinghua.edu.cn, }}\\ {\tt\small{jiashi@andrew.cmu.edu, guangliangcheng2014@gmail.com}}
}

\maketitle

\begin{abstract}
We focus on bridging domain discrepancy in lane detection among different scenarios to greatly reduce extra annotation and re-training costs for autonomous driving. Critical factors hinder the performance improvement of cross-domain lane detection that conventional methods only focus on pixel-wise loss while ignoring shape and position priors of lanes. To address the issue, we propose the Multi-level Domain Adaptation (MLDA) framework, a new perspective to handle cross-domain lane detection at three complementary semantic levels of pixel, instance and category. Specifically, at pixel level, we propose to apply cross-class confidence constraints in self-training to tackle the imbalanced confidence distribution of lane and background. At instance level, we go beyond pixels to treat segmented lanes as instances and facilitate discriminative features in target domain with triplet learning, which effectively rebuilds the semantic context of lanes and contributes to alleviating the feature confusion. At category level, we propose an adaptive inter-domain embedding module to utilize the position prior of lanes during adaptation. In two challenging datasets, i.e. TuSimple and CULane, our approach improves lane detection performance by a large margin with gains of 8.8\% on accuracy and 7.4\% on F1-score respectively, compared with state-of-the-art domain adaptation algorithms.
\end{abstract}

\begin{figure}[ht]
  \centering
  \includegraphics[width=0.9\linewidth]{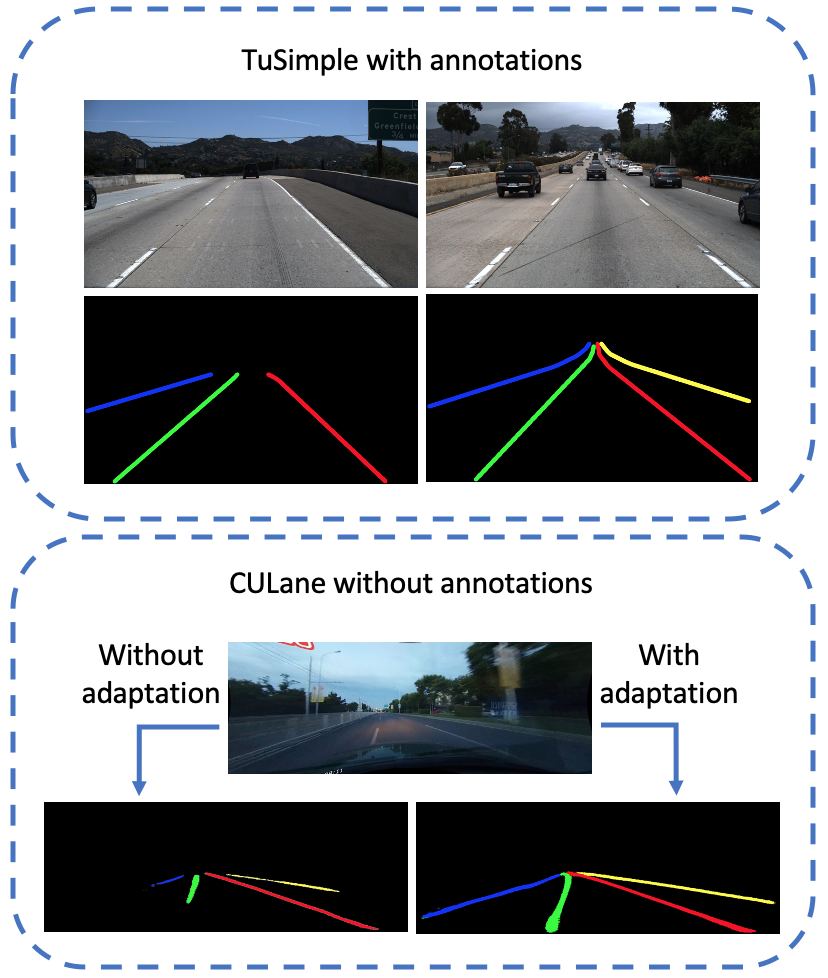}
  \caption{
  Lane detection results improve significantly on CULane \cite{pan2018spatial} with our proposed unsupervised domain adaptation approaches (bottom). The source domain model is trained on TuSimple \cite{TuSimple} with annotations (top).}
  \label{tab:fig1}
\vspace{-0.4cm}
\end{figure}

\begin{figure*}[ht]
  \centering
  \includegraphics[width=\linewidth]{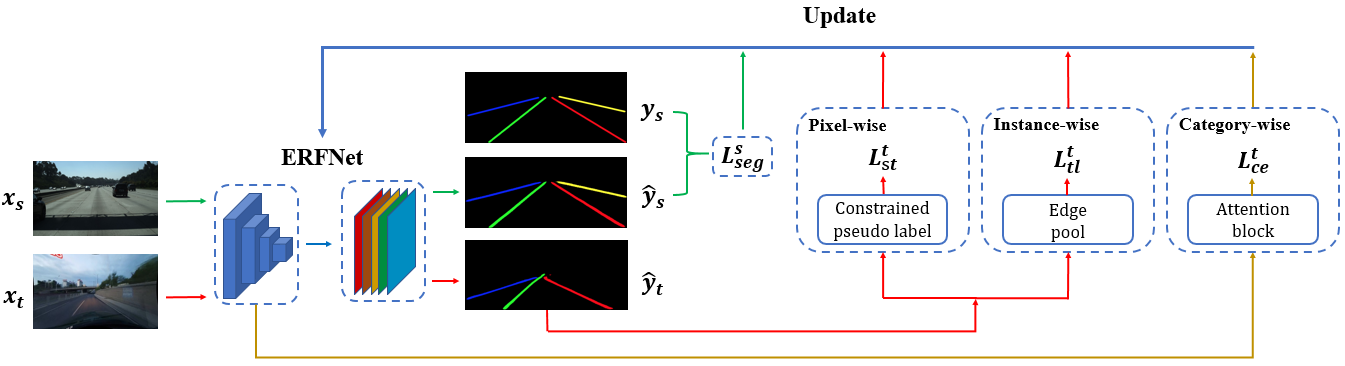}
  \caption{
  \textbf{Approach overview (MLDA).} We use ERFNet \cite{romera2017erfnet} as backbone. $x_{s}$ and $y_{s}$ refer to the input image and groundtruth in source domain respectively. $x_{t}$ is the input image in target domain. $\hat{y_{s}}$ and $\hat{y_{t}}$ are segmentation results of $x_{s}$ and $x_{t}$ respectively.}
  \label{tab:fig2}
\vspace{-0.4cm}
\end{figure*}

\section{Introduction}
\footnote{Proceedings of the CVPR 2022 Workshop of Autonomous Driving}Lane detection \cite{hillel2014recent,aly2008real,kim2008robust} is a key component for camera-based perception in autonomous vehicles which is widely applied in modules such as lane keeping assist (LKA) and lane departure warning (LDW) \cite{narote2018review,jung2013efficient}. Benefit from the rapid development of deep neural networks, lane detection approaches have made tremendous advances ~\cite{hou2019learning,li2019line,pan2018spatial,neven2018towards,garnett20193d,ghafoorian2018gan}. For safety and efficiency considerations, the lane detection system is required to exhibit high stability and accuracy in various challenging environments. For example, under severe weather conditions, vehicles need to recognize lanes accurately to perform correct path planning and decision making. However, when models trained on a specific scene (source domain) are directly applied to another (target domain), the performance may drop dramatically because of the domain shift \cite{pan2009survey, wang2018deep}. Although re-labeling and re-training can bring some improvements, they take high labeling costs and large time consumption~\cite{richter2016playing}.

In the past few years, deep neural networks have shown great potential in semantic segmentation  \cite{long2015fully,chen2017deeplab,chen2017rethinking,zhao2017pyramid,romera2017erfnet,chen2018encoder}, among which representative datasets are PASCAL VOC \cite{everingham2015pascal}, Cityscapes \cite{cordts2016cityscapes}, etc. Current lane detection works based on semantic segmentation networks \cite{hou2019learning,pan2018spatial,neven2018towards} mainly focus on improving accuracy in one particular dataset, e.g. TuSimple \cite{TuSimple} or CULane \cite{pan2018spatial}, assuming that the training and testing data have 
the same distribution. We endeavor to improve cross-dataset performance of lane detection models by unsupervised domain adaptation (UDA) using unlabeled data, as illustrated in Figure~\ref{tab:fig1}.

Unsupervised domain adaptation is a research field which aims to learn well-performed models in a target domain without training labels. 
Semi-supervised learning \cite{lee2013pseudo,hong2018conditional,grandvalet2005semi} approaches are applied to UDA, in which representative methods are entropy minimization \cite{yan2017mind} and self-training \cite{laine2016temporal,tarvainen2017mean}.
Adversarial approaches have been explored in \cite{ganin2014unsupervised,tzeng2017adversarial,long2016unsupervised,vu2019advent}. CBST \cite{zou2018unsupervised} generates pseudo-labels to 
minimize cross-entropy loss in target domain. Yan \textit{et al.} \cite{yan2017mind} minimizes the distance between source and target domains by maximum mean discrepancies (MMD). In addition, there are generation networks that make features domain-invariant \cite{hoffman2017cycada,sankaranarayanan2018learning,wu2018dcan,wu2019ace}.

For lane detection adaptation, directly applying the above methods may obtain high false positive or false negative rates. We summarize the reasons for performance drops into two factors. Firstly, the imbalanced proportion of lane and background classes and the different appearance of lanes between source and target domain bring discrepancy in class confidence, thus in self-training low-entropy class (background) is always easy to be learned and high-entropy class (lane) tends to be suppressed. Secondly, pixel-wise loss functions in UDA fail to grasp the shape and position priors of lanes, which may lead to disconnected lane predictions and class confusion.

Therefore, we propose a Multi-level Domain Adaptation framework in domain adaptation of lane detection. At pixel level, to balance class confidence of background and non-background (lane), we modify the self-training strategy in previous method \cite{french2017self} to a confidence constrained manner, which keeps the pseudo labels avoid of being dominated by the background class. At instance level, we use triplet learning with edge pooling in target domain to make feature refinements, that is, we train the network to learn discriminative feature embeddings of lane and background class by optimizing a triplet loss function, which will increase the distance of feature embeddings in different classes by a large margin, and pull the feature embeddings of lane instances closer in the embedding space, thus reducing feature confusion. By edge pooling, the disconnected parts of lanes in target domain are adaptively expanded in four directions of up, down, left, and right to enhance lane edges and get fine triplet masks which can better fit the lane areas. At category level, we integrate the existence of lanes with self-training and propose an adaptive inter-domain embedding module to utilize the position prior of lanes. Our contributions are summarized as follows:
\begin{itemize}
\item {We propose a multi-level domain adaptation approach for lane detection. In pixel-level adaptation, we propose a cross-domain class balance scheme based on confidence constrained self-training. Beyond the pixels, we present instance-level adaptation by triplet learning with the novel edge pooling to achieve discriminative features between lane instances and background, which can greatly reduce the discontinuity of lanes. Moreover, we utilize the position prior of lanes with category-level adaptation by proposing an adaptive inter-domain embedding module and integrating the existence of lanes with self-training.}
\item {
Our approach provides a strong baseline in the field of cross-domain lane detection, which surpasses the state-of-the-art domain adaptation algorithms by a large margin with gains of 8.8\% on accuracy and 7.4\% on F1-score in two challenging lane detection datasets respectively.}
\end{itemize}

\begin{figure*}[ht]
  \centering
  \includegraphics[width=0.22\linewidth]{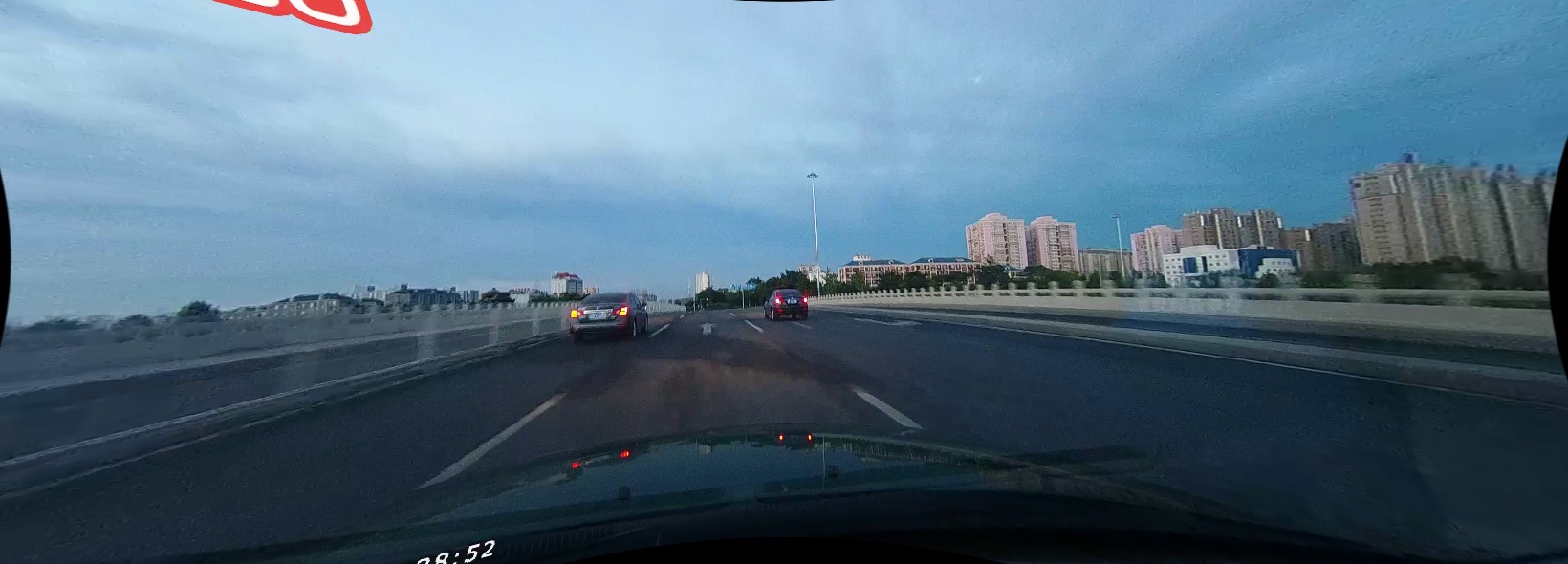}
  \includegraphics[width=0.22\linewidth]{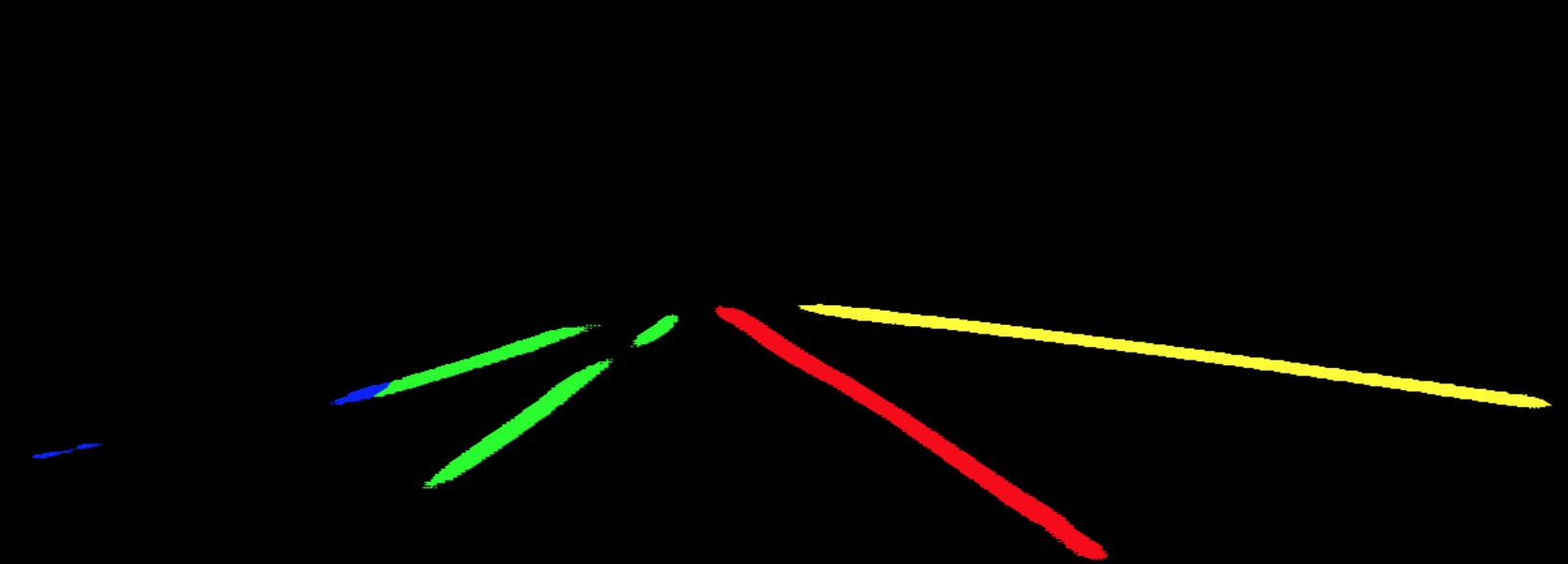}
  \includegraphics[width=0.22\linewidth]{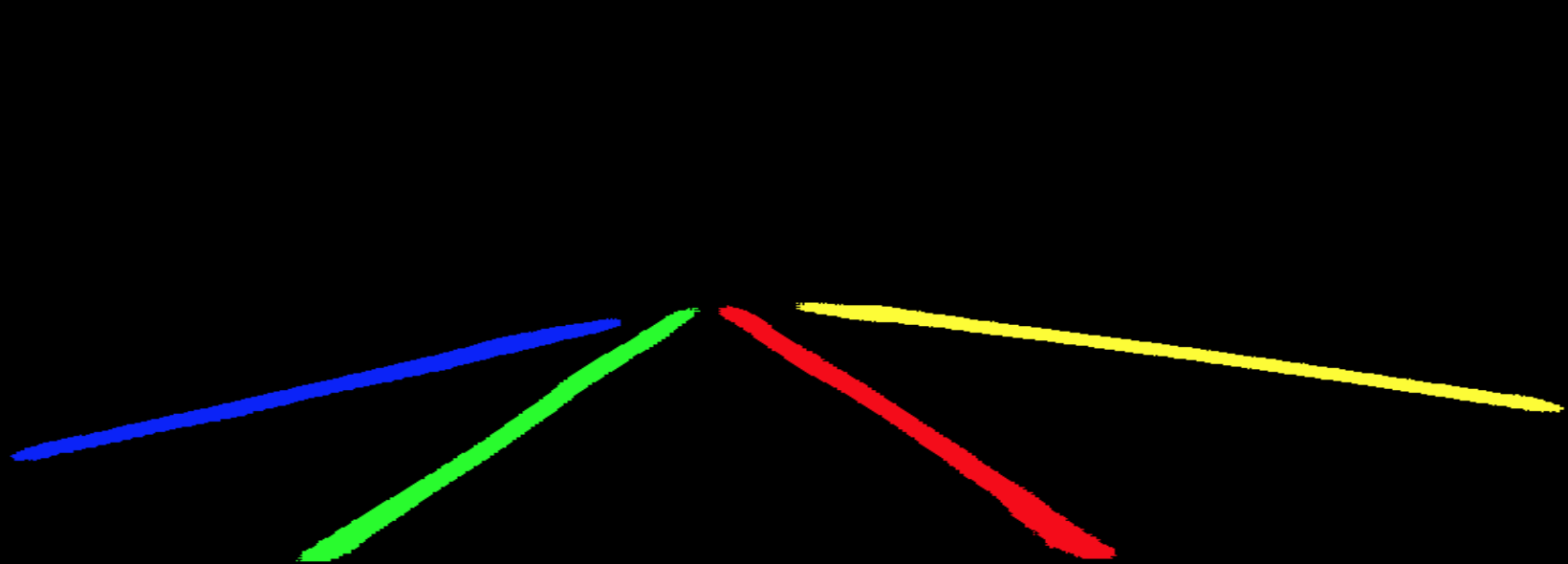}
  \includegraphics[width=0.22\linewidth]{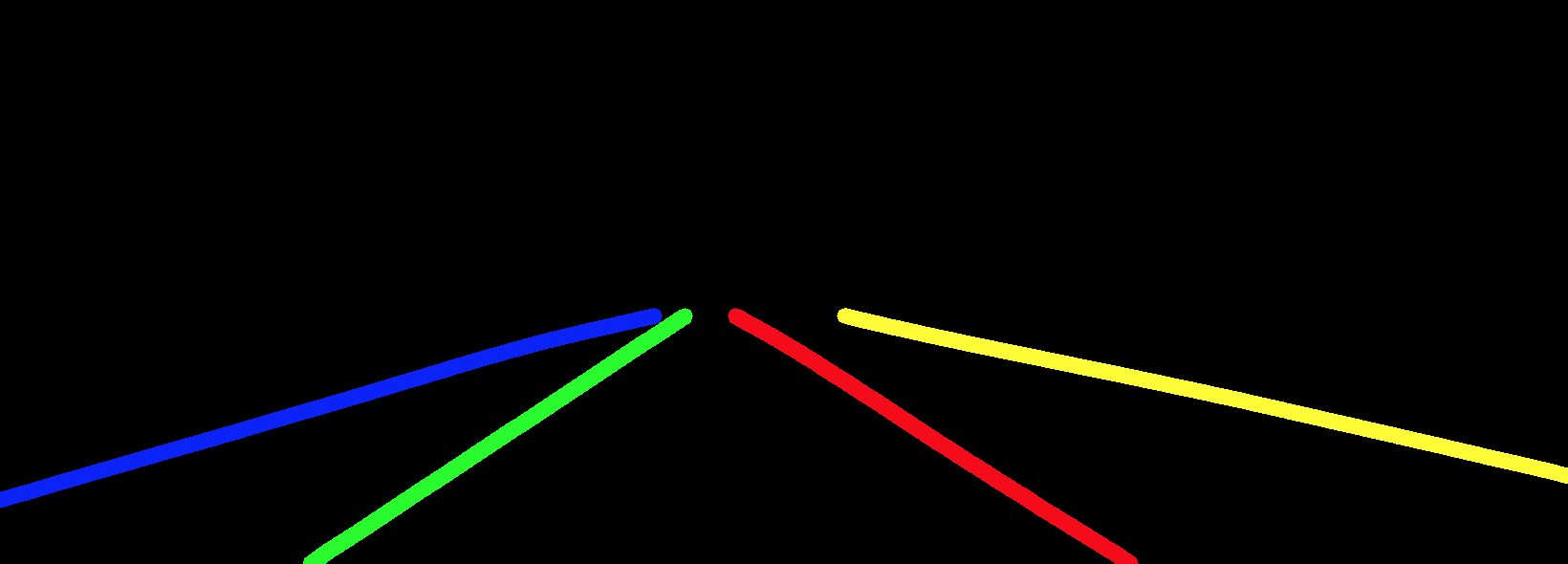}
  \includegraphics[width=0.22\linewidth]{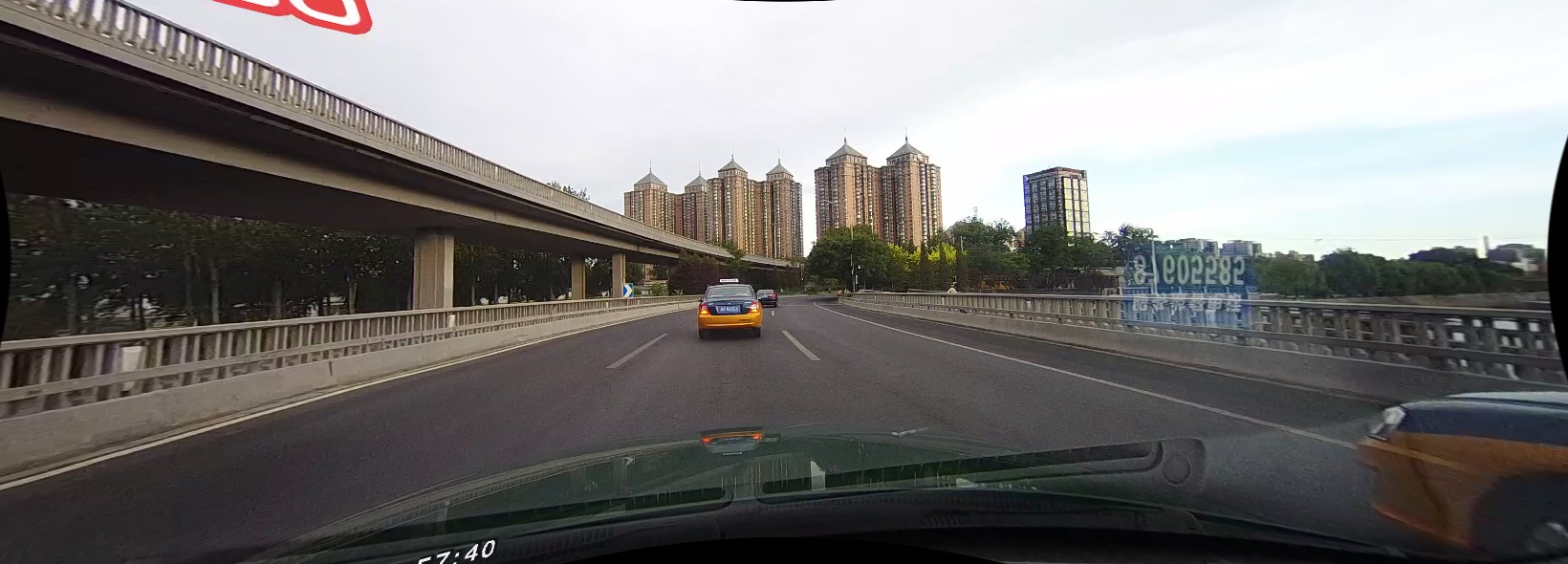}
  \includegraphics[width=0.22\linewidth]{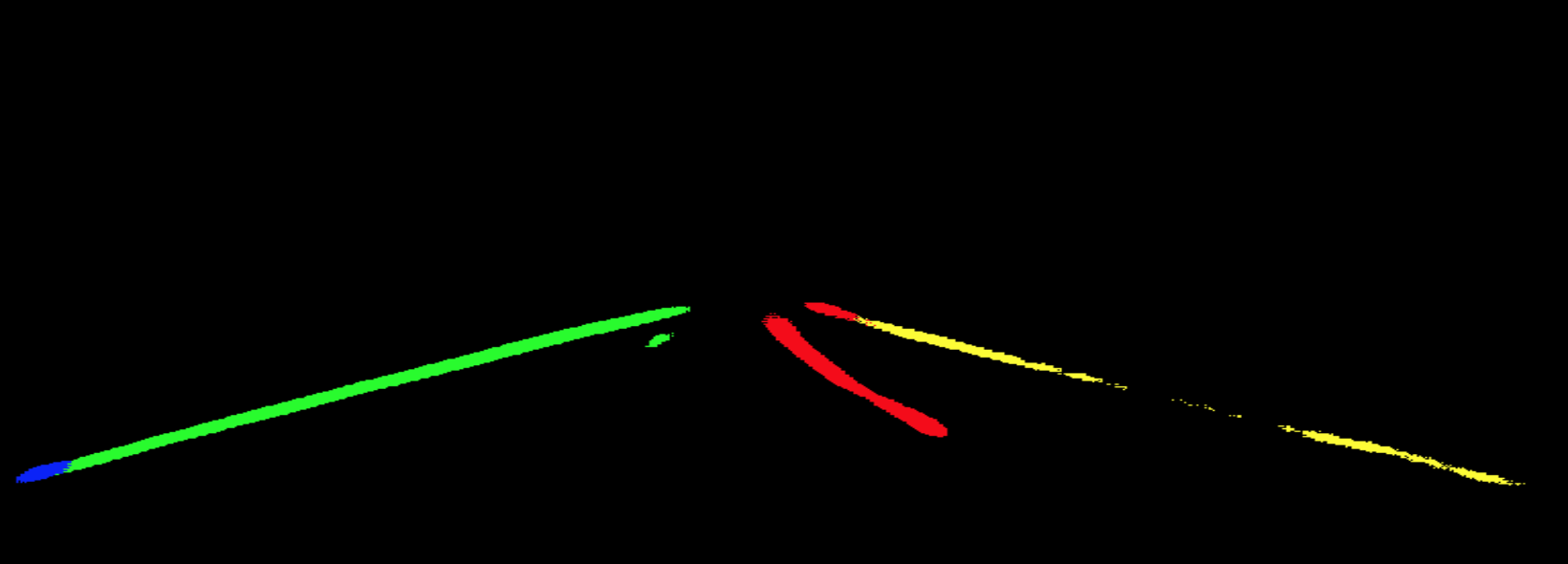}
  \includegraphics[width=0.22\linewidth]{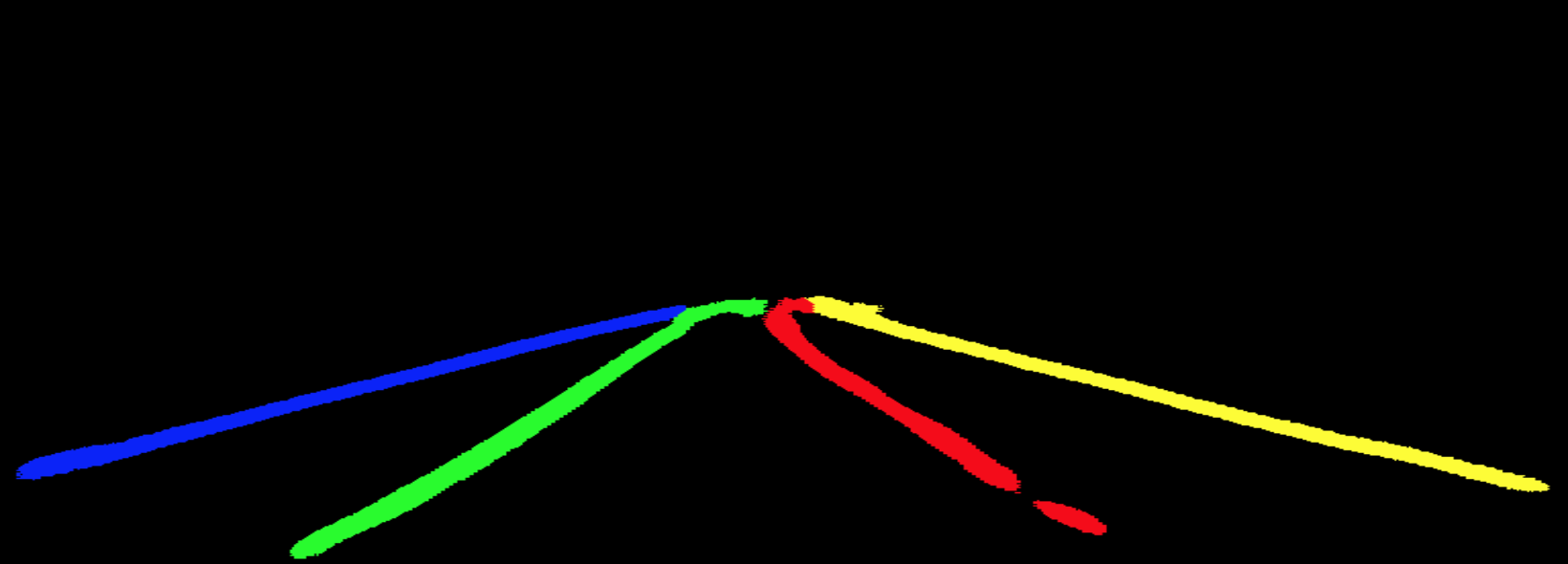}
  \includegraphics[width=0.22\linewidth]{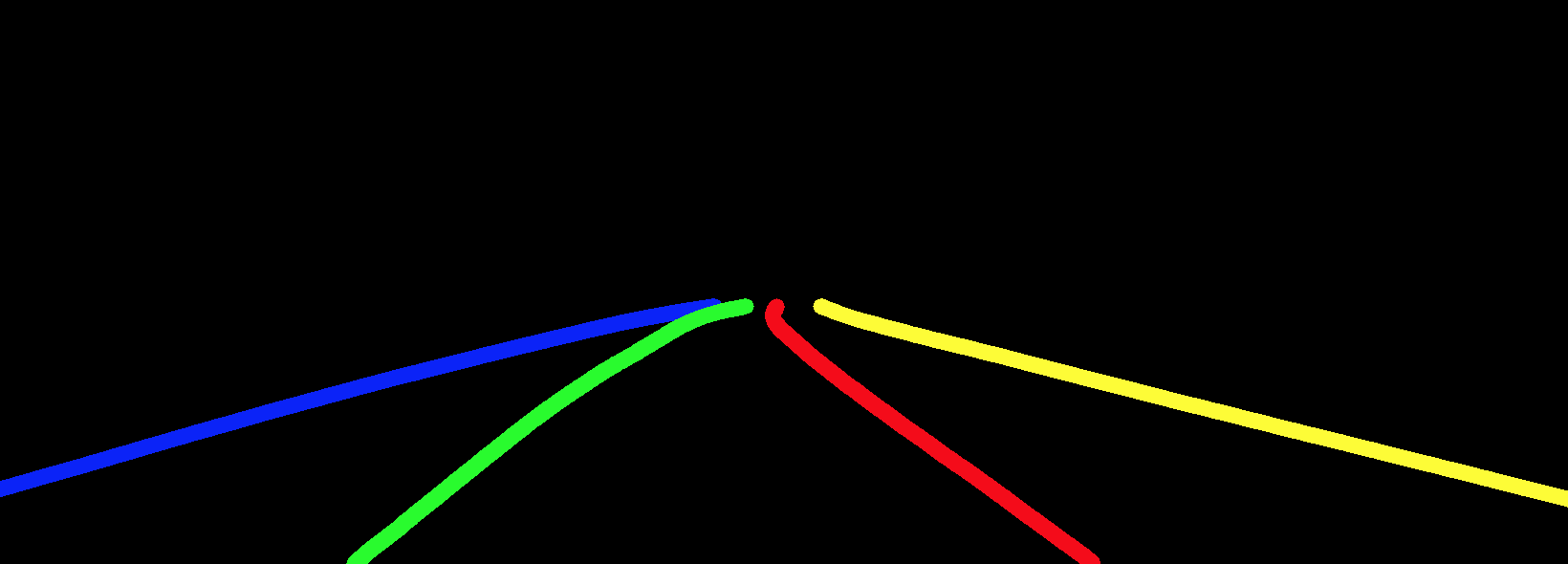}
  \includegraphics[width=0.22\linewidth]{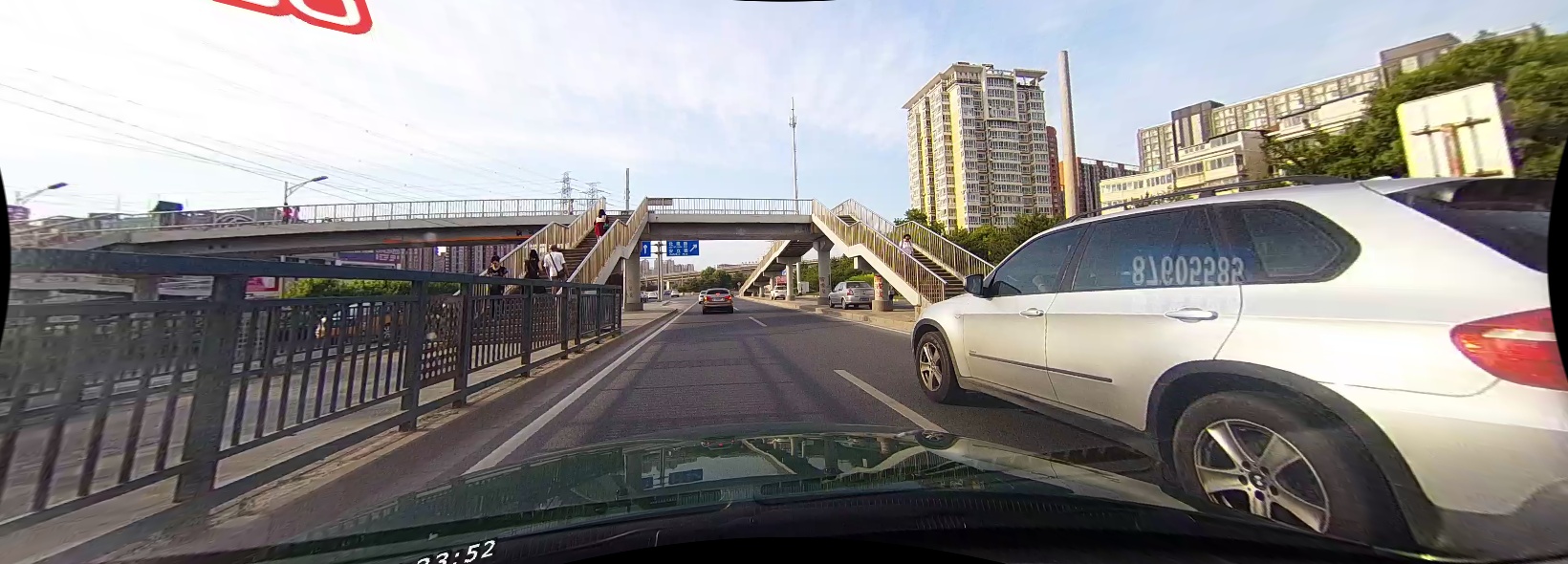}
  \includegraphics[width=0.22\linewidth]{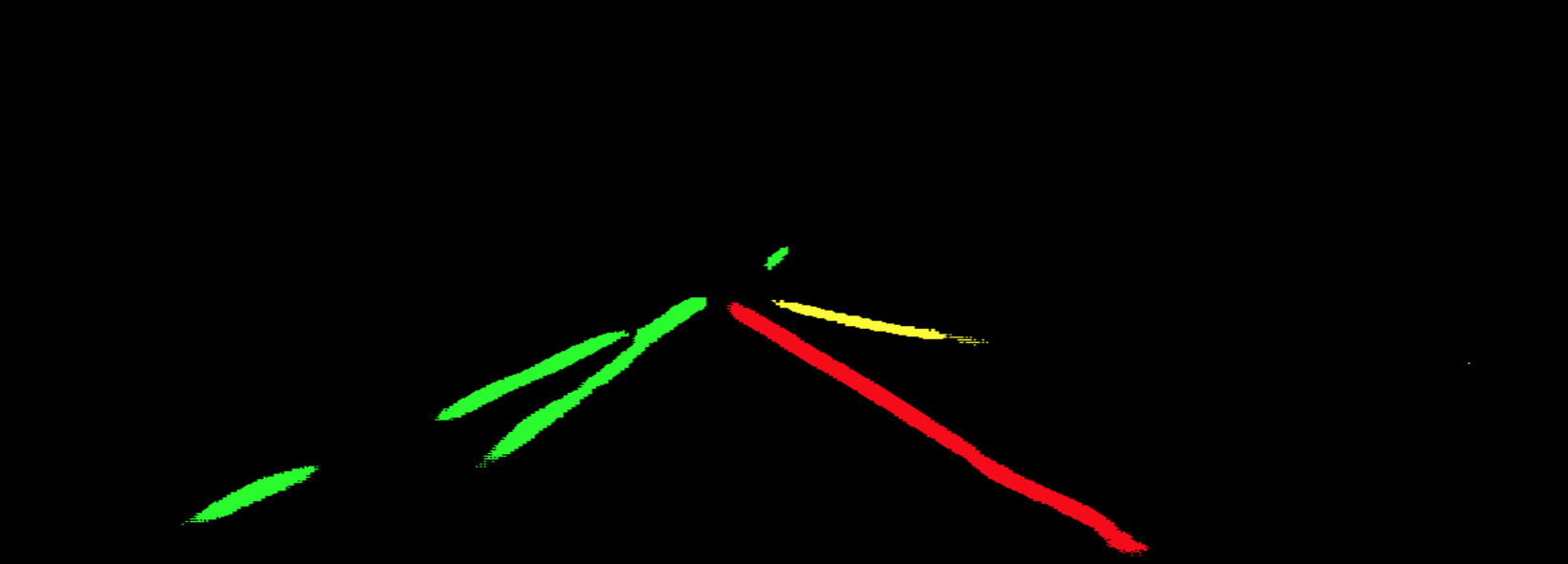}
  \includegraphics[width=0.22\linewidth]{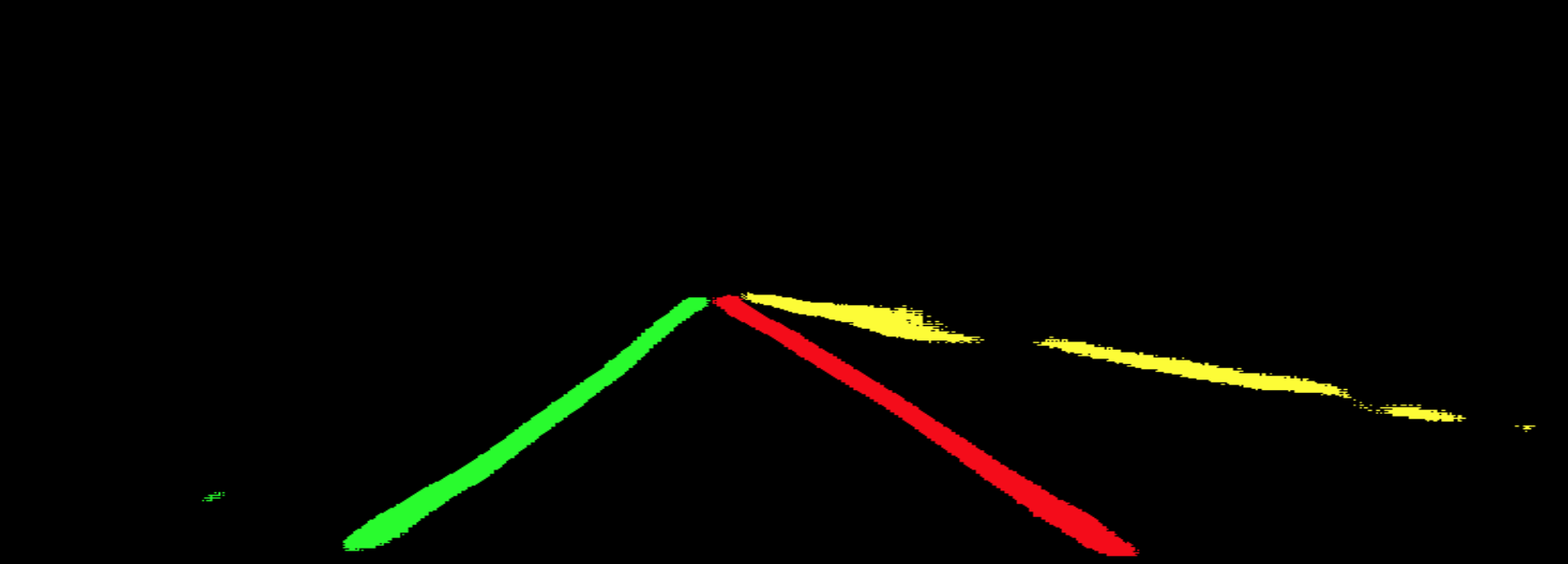}
  \includegraphics[width=0.22\linewidth]{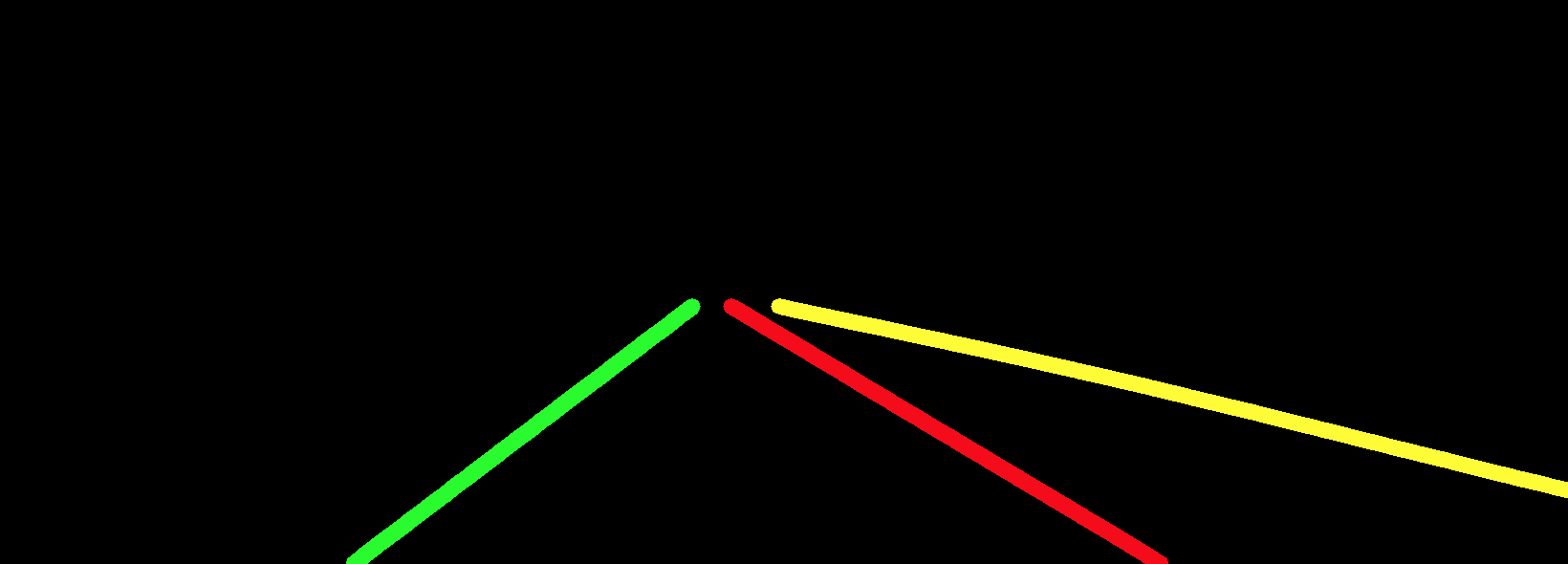}
  \caption{\textbf{Visualization results on TuSimple $\rightarrow$ CULane}. From left to right are (a) Input image, (b) Source only, (c) Ours (MLDA) and (d) Groundtruth.}
  \label{tab:fig3}
\vspace{-0.4cm}
\end{figure*}

\begin{figure*}[ht]
  \centering
  \includegraphics[width=0.22\linewidth]{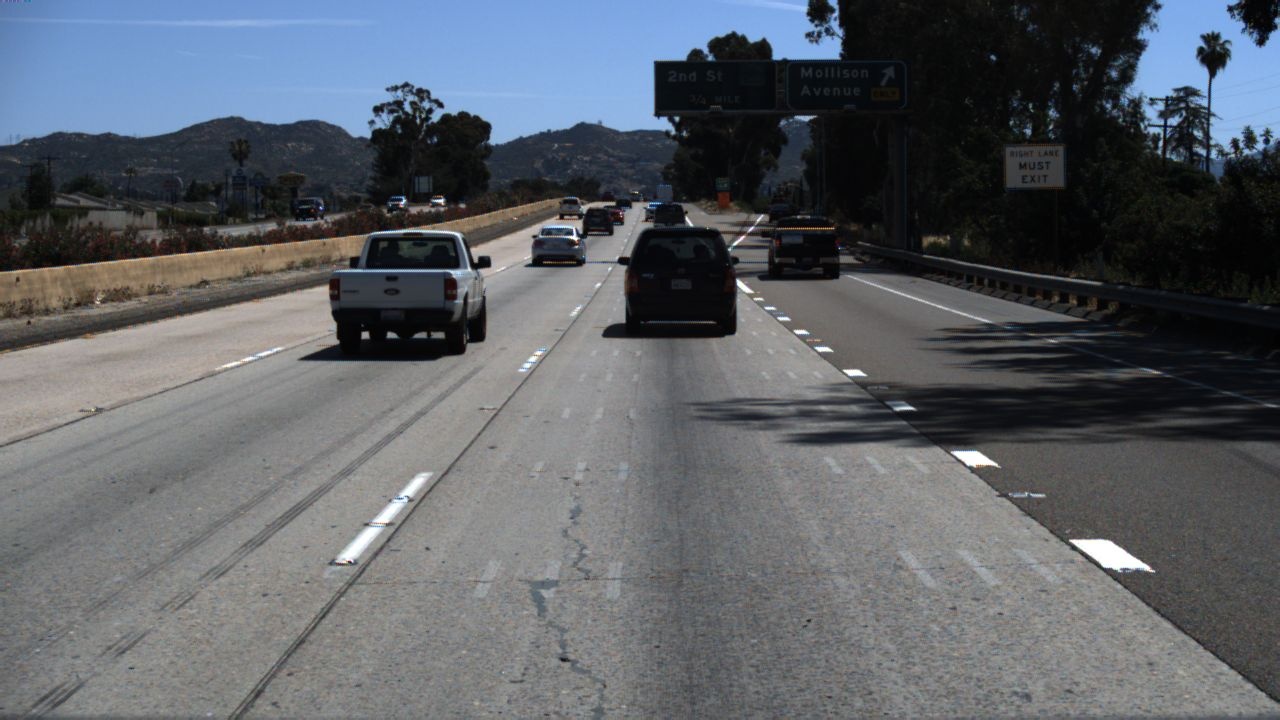}
  \includegraphics[width=0.22\linewidth]{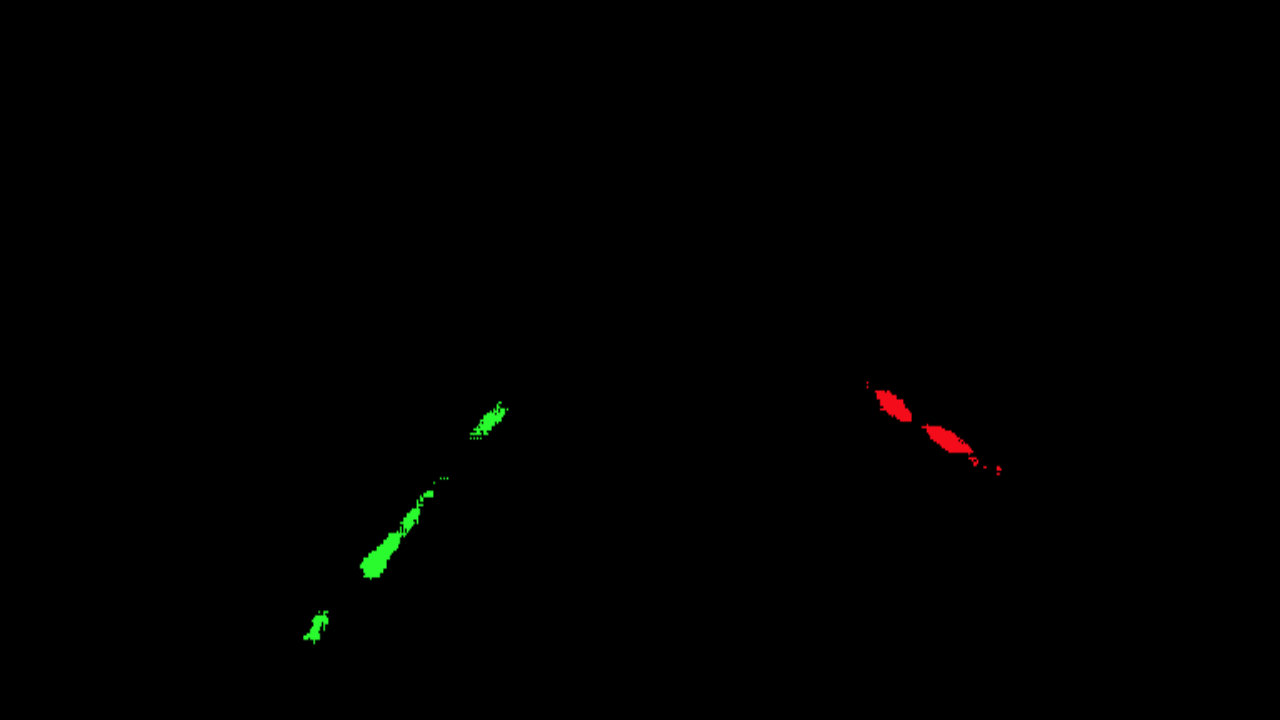}
  \includegraphics[width=0.22\linewidth]{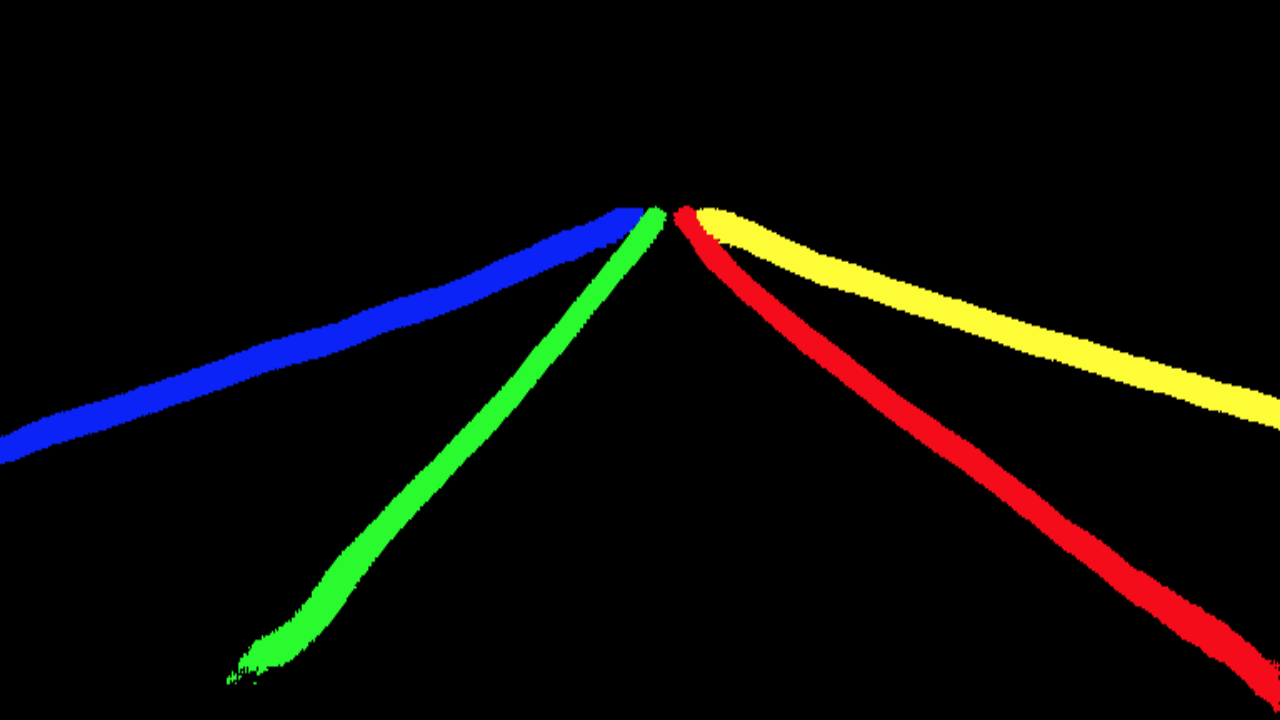}
  \includegraphics[width=0.22\linewidth]{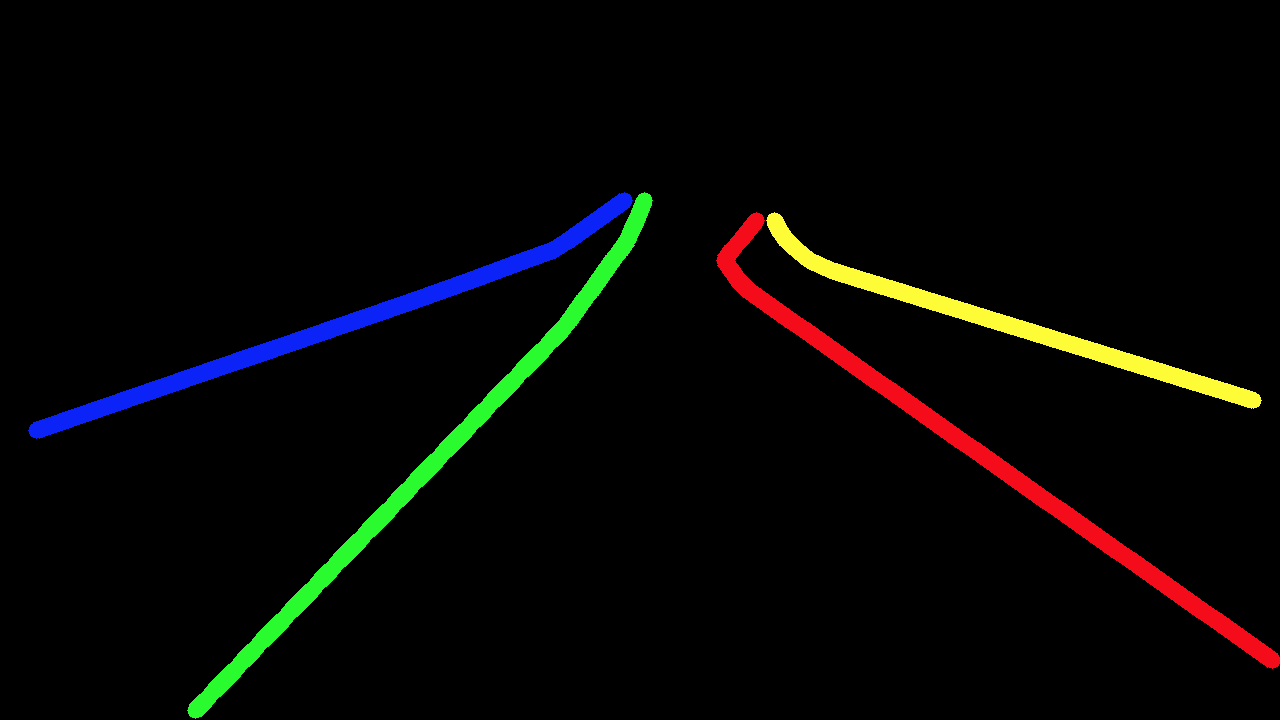} 
  \includegraphics[width=0.22\linewidth]{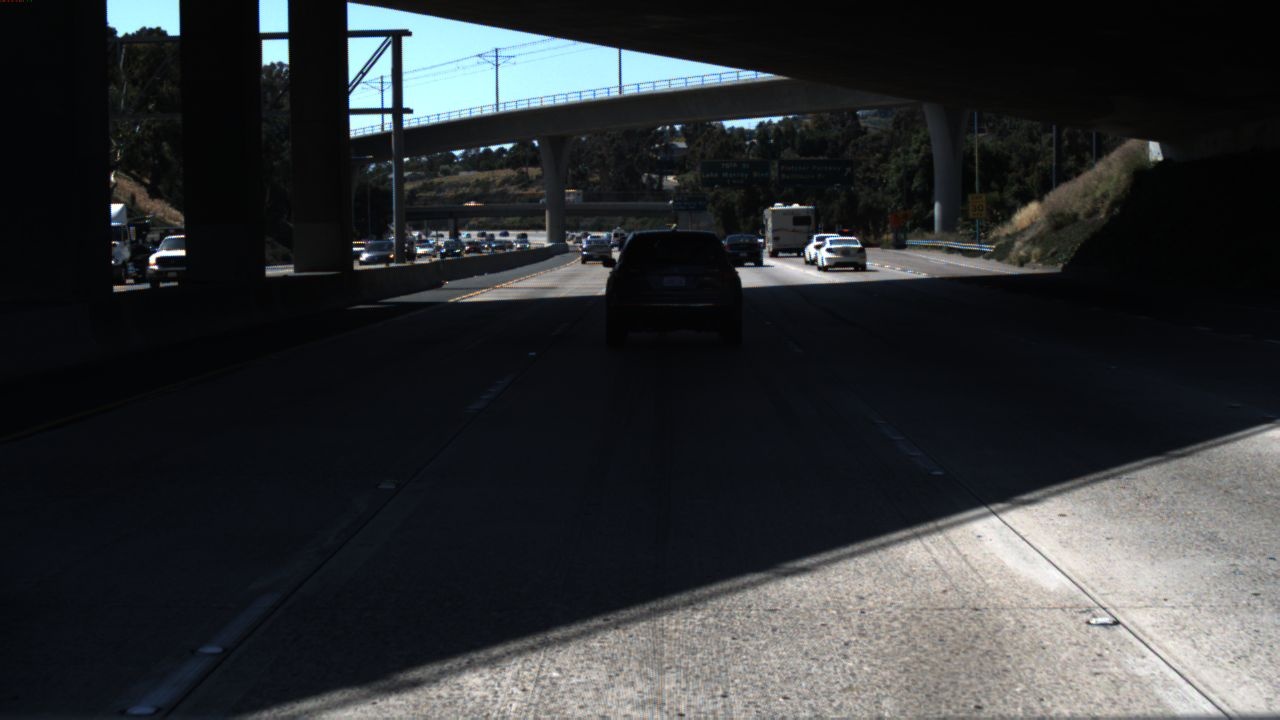}
  \includegraphics[width=0.22\linewidth]{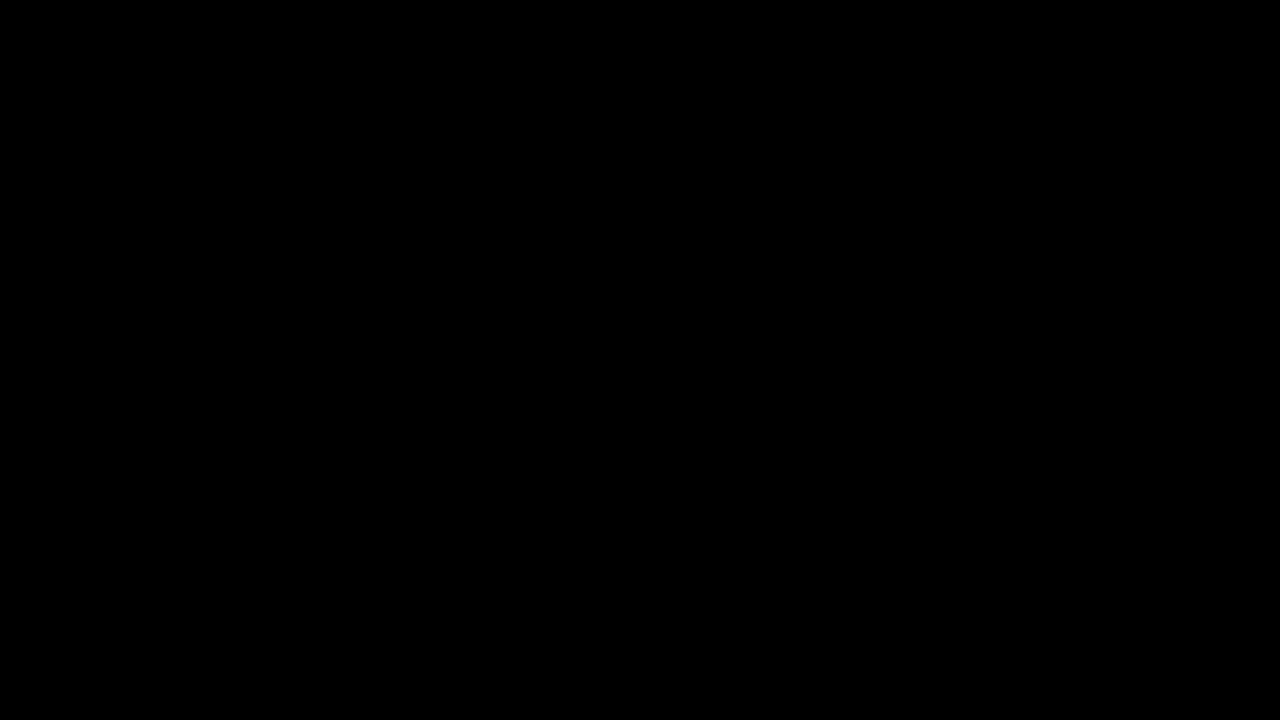}
  \includegraphics[width=0.22\linewidth]{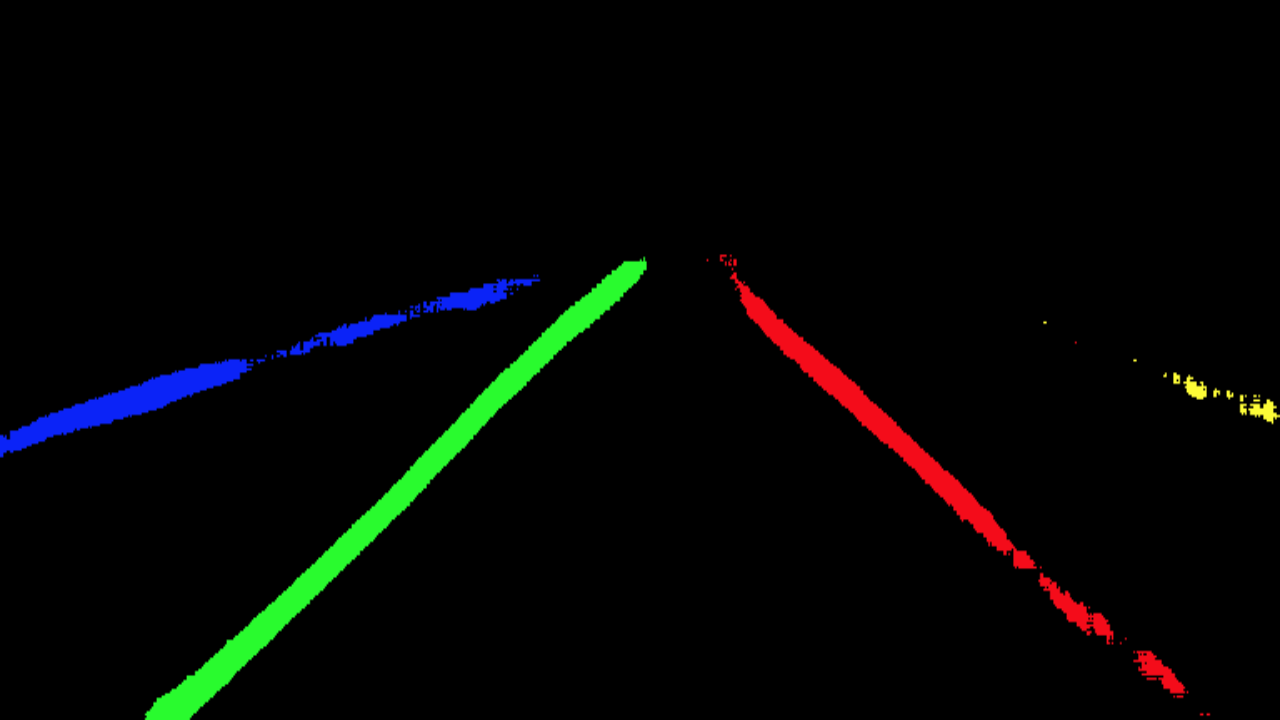}
  \includegraphics[width=0.22\linewidth]{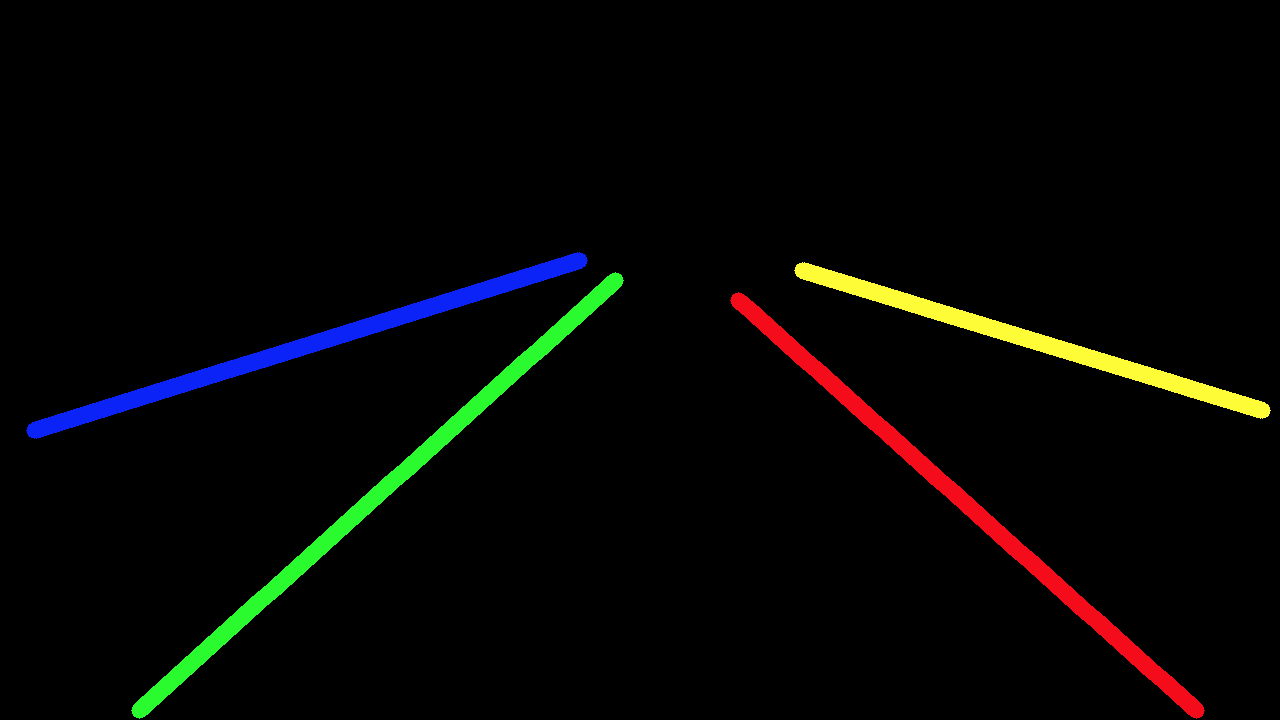}
  \includegraphics[width=0.22\linewidth]{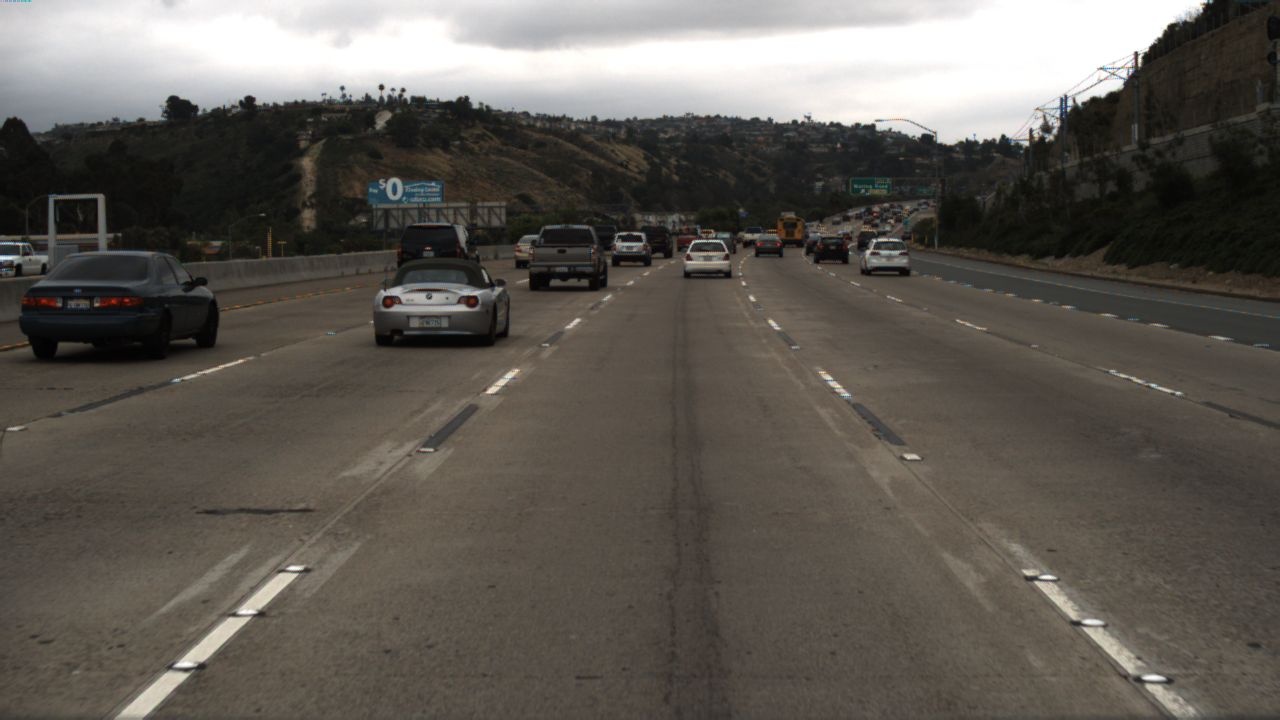}
  \includegraphics[width=0.22\linewidth]{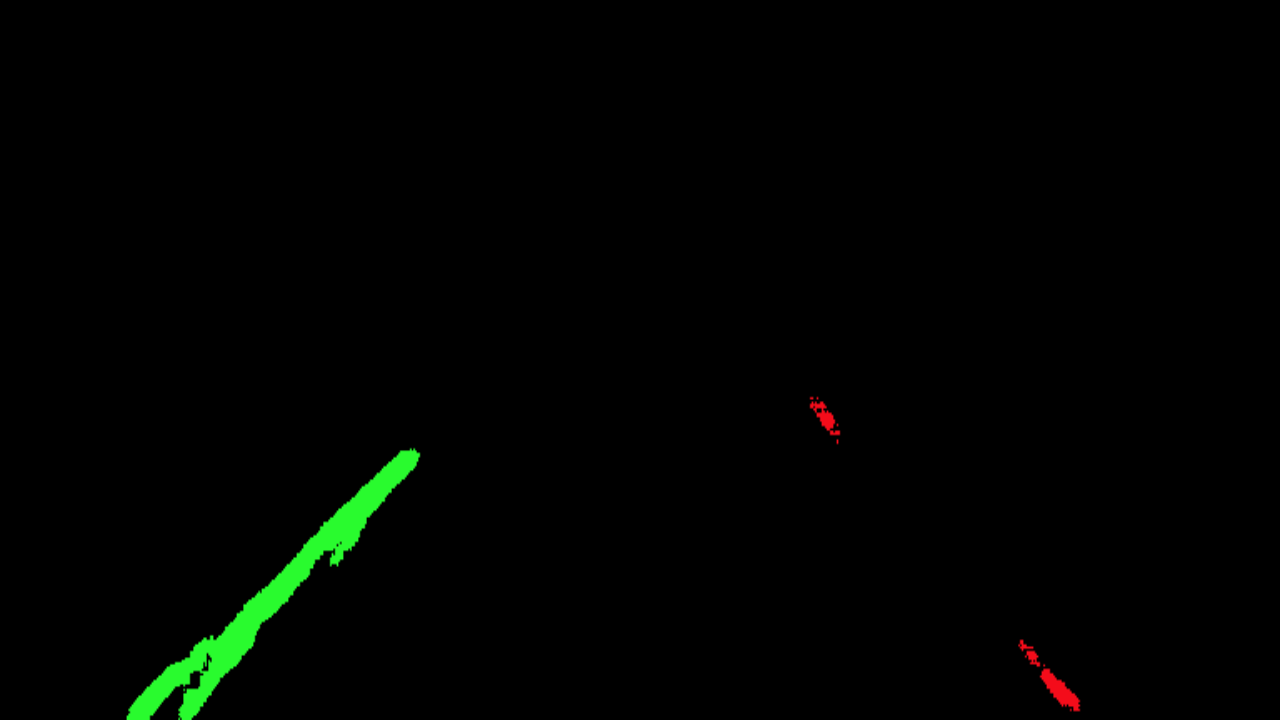}
  \includegraphics[width=0.22\linewidth]{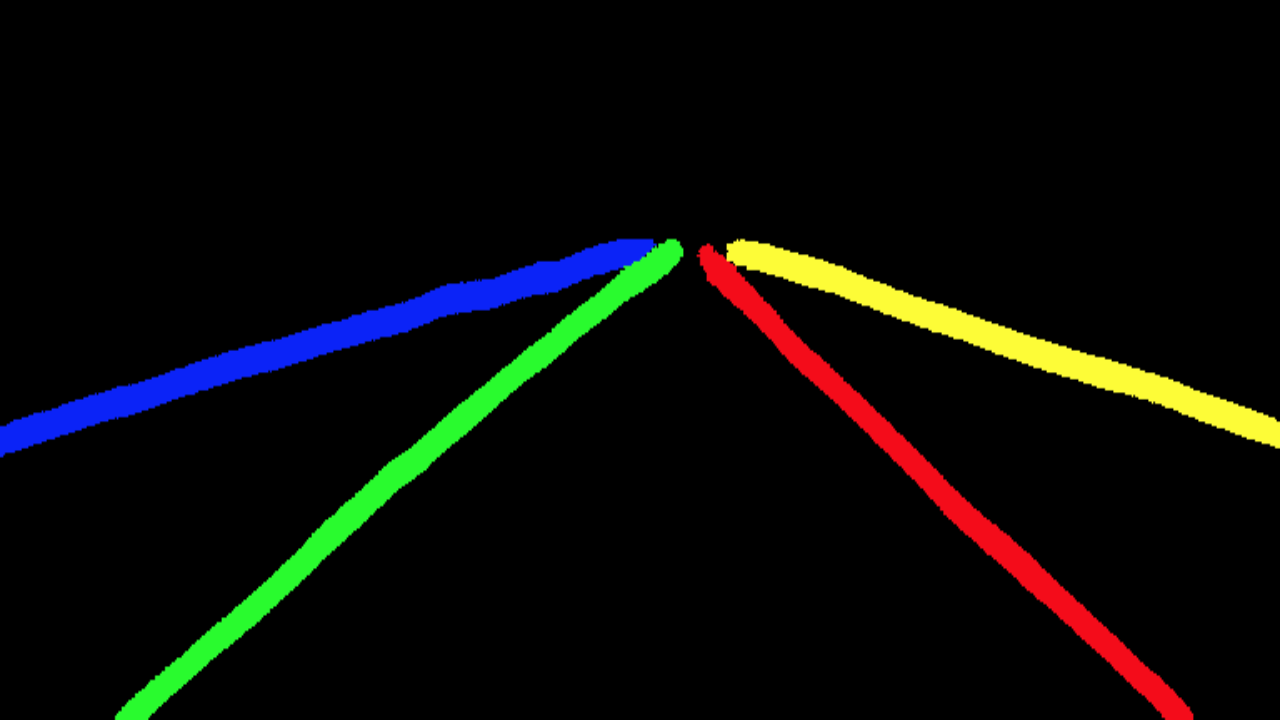}
  \includegraphics[width=0.22\linewidth]{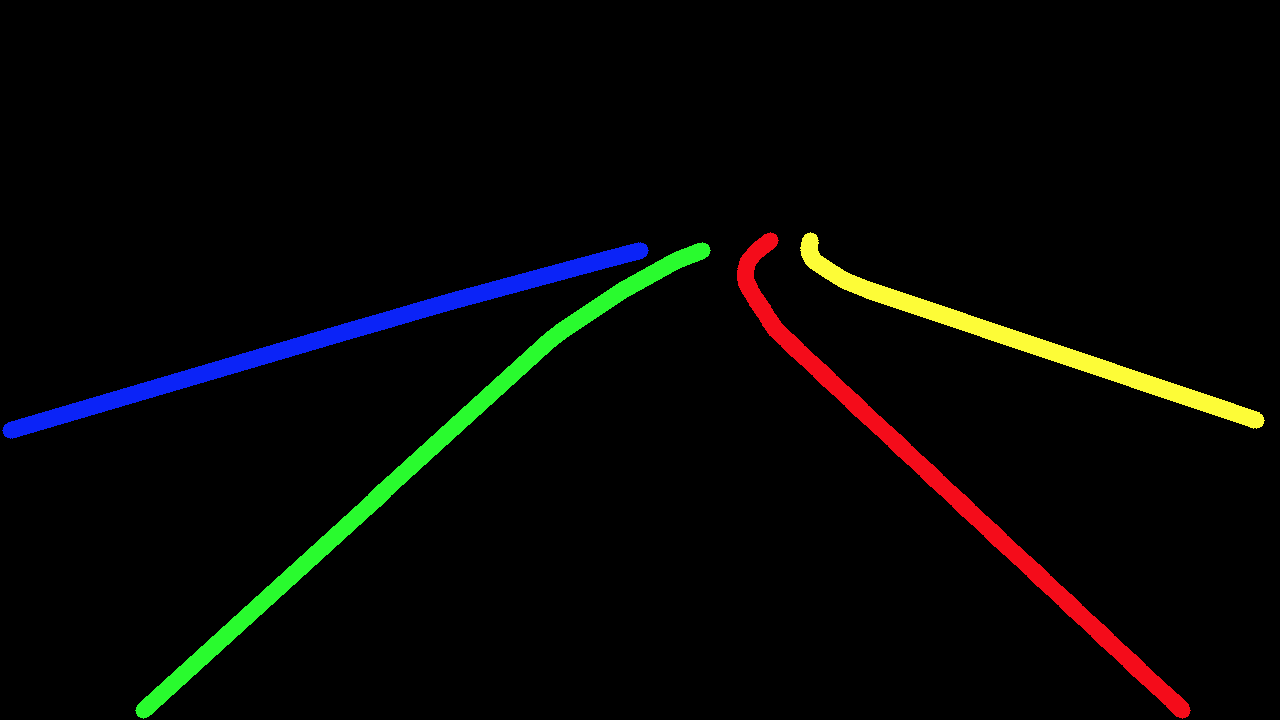}
  \caption{\textbf{Visualization results on CULane $\rightarrow$ TuSimple}. From left to right are (a) Input image, (b) Source only, (c) Ours (MLDA) and (d) Groundtruth.}
  \label{tab:fig3-1}
\vspace{-0.4cm}
\end{figure*}

\section{Related Work}
\textbf{Semantic segmentation on lane detection.} Semantic segmentation is the task of assigning pixel-level tags to images. After many years of development, semantic segmentation models based on deep neural networks achieve great success. In practice, segmentation models are widely used in lane detection for self-driving cars. Pan \textit{et al.} \cite{pan2018spatial} propose a spatial encoder on the four directions to encode the context information of lanes and obtain strong results on the TuSimple Lane Detection Challenge \cite{TuSimple}.  Neven \textit{et al.} \cite{neven2018towards} design a embedding branch to cluster lane instances from binary segmentation results. Hou \textit{et al.} \cite{hou2019learning} introduce self attention distillation applied to different network architecture for lane detection. Li \textit{et al.} \cite{li2019line} propose the line proposal unit (LPU) to locate accurate traffic curves. However, the performance requires a high-quality labeled dataset and labeling for each new scene brings extra cost in both time and human labor. In order to reduce the manual labeling workload and re-training cost, the problem of domain discrepancy needs to be solved to keep models trained from a labeled dataset to get similar performance in another with the absence of annotations.

\textbf{Domain adaptation.} In order to solve the domain discrepancy problem, we focus on unsupervised domain adaptation technology which is a hot topic in the research of classification, segmentation and detection tasks. Domain adaptation uses data without groundtruth to improve performance for actual tasks. When the feature distribution between training and test data is different, the performance of the model would suffer significantly drops. The core idea of unsupervised domain adaptation is to learn the domain-invariant features, which means minimizing the difference of feature distribution between source and target domains. The approaches include self-training, adversarial network and generating methods.

\begin{figure*}[ht]
  \centering
  \includegraphics[width=\linewidth]{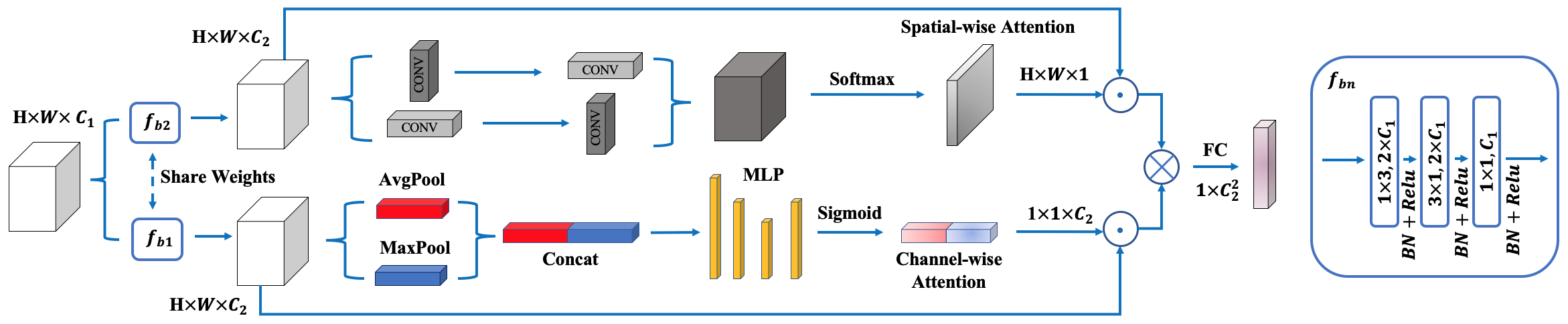}
  \caption{
  \textbf{Adaptive inter-domain embedding module.} The multi-layer perceptron (MLP) includes three layers (FC + Relu). Three FC layers squeeze feature maps to 1/2, 1/16 and 16 times respectively.}
  \label{tab:fig4}
\vspace{-0.4cm}
\end{figure*}

Self-training utilizes the prediction of the previous state model as pseudo labels in the target domain to assist the training of the current model. In earlier works, self-training is mainly used in semi-supervised learning methods, such as \cite{laine2016temporal,tarvainen2017mean,french2017self}. Entropy minimization that encourages minimization of cluster allocation \cite{grandvalet2005semi} is one of the most popular methods in semi-supervised learning. Zou \textit{et al.} \cite{zou2018unsupervised} extend self-learning to semantic segmentation based on class balance and spatial priors. Lian \textit{et al.} \cite{lian2019constructing} propose that self-training is also a kind of curriculum-style domain adaptation method, and combine CBST \cite{zou2018unsupervised} with CDA \cite{zhang2019curriculum} to achieve state-of-the-art performance. However, the gradient of the entropy inclines to samples that are easy to transfer. Chen \textit{et al.} \cite{chen2019domain} balance the gradient of well-classified target samples and makes difficult samples be trained more efficiently.

Another method in UDA is adversarial training, which involves two networks including segmentation and discriminator. The discriminator needs to obtain the feature map from the segmentation network and predict the domain of input, which deceives each other to make the features from the two domains to reach a similar distribution. Hoffman \textit{et al.} \cite{hoffman2016fcns} first adopt adversarial methods for semantic segmentation in unsupervised domain adaptation and adapt the label statistics from the source domain to obtain a specific category. Chen \textit{et al.} \cite{chen2017no} propose global and class alignment, where class alignment is obtained by the adversarial training of the network. Hong \textit{et al.} \cite{hong2018conditional} use a residual network to make source and target feature maps similar in adversarial training. Vu \textit{et al.} \cite{vu2019advent} combine adversarial training with minimization of the prediction entropy of target samples to achieve state-of-the-art performance.

Besides self-training and adversarial methods, another idea is to transfer styles between images from different domains. CyCADA \cite{hoffman2017cycada} utilizes CycleGAN \cite{isola2017image} to construct a labeled dataset, which is similar to the target dataset. In DCAN \cite{wu2018dcan}, two networks make channel-by-channel feature alignment to learn a segmentation network on the generated images. The generated images have the content of source and the style of target, and the source segmentation maps can be used as groundtruth.

\section{Method}
In this section, we focus on balancing the relationship between the confidence of background and non-background classes and leveraging the shape and position priors of lanes in domain adaptation. 
Thus, we propose our Multi-level Domain Adaptation method (MLDA) including constrained self-training at pixel level, triplet learning with edge pooling at instance level and adaptive inter-domain embedding module at category level. Figure~\ref{tab:fig2} is an overview of our approach.

\subsection{Self-training with confidence constraints}
The distribution alignment is critical for the domain adaptation performance. Models trained on source domain tend to produce low-entropy predictions on source images, but high-entropy predictions on target ones \cite{grandvalet2005semi}. Specifically for the lane detection task, the proportion of line pixels (foreground class) are relatively small and the lines have different position prior in the target domain, however the road and surrounding pixels (background class) are relatively stable and easy to be adapted. This characteristic leads to an imbalanced distribution where the background pixels always have much higher probabilities than the line pixels in the target domain. As a result, the background class pixels would be dominant in the generated pseudo labels so that the model predicts only background-class samples during the process of self-training.

In order to solve this problem, we propose a confidence constraint strategy for self-training, which makes it achievable for the model to be aware of the intrinsic distribution in each class when producing the pseudo labels. Under the control of probability gate, confidence imbalance between background and non-background classes is alleviated, and much more accurate pseudo labels are generated by iterative network learning.

Self-training \cite{zou2018unsupervised} generates pseudo labels for images in the target domain. It is solved by alternating optimization based on the following steps:

1) Infer the values of the target labels

2) Update weights of the network

\begin{equation}
\begin{split}
L_{st} & = L^{s}_{seg} + L^{t}_{st}
\\ & = min \sum_{s \in S}^{ } CE(y_{s},\hat{y_{s}}) +
\lambda_{st} \sum_{t \in T}^{ } CE(y_{t},\hat{y_{t}})
\end{split}
\label{tab:eq1}
\end{equation}

where $\hat{y_{s}}$ and $\hat{y_{t}}$ are segmentation results of source and target images respectively. $y_{s}$ is groundtruth of one source image and $y_{t}$ is pseudo label of one target image. The first term and second term sum up pixel-wise cross-entropy losses over the source domain images $(s \in S)$ and target domain images $(t \in T)$ respectively.

In lane detection adaptation, we modify the method in generating pseudo labels of target images in step 1) to a relatively simple but effective way considering the imbalanced confidence between classes. In this step, we propose probability constrained self-training, in which the thresholds $\alpha_{c}$ are set for background and non-background classes respectively to generate pseudo labels effectively:

\begin{equation}
\begin{split}
y_{t}(i,j) = \left\{\begin{matrix}
argmax_{C} \hat{y_{t}}(i,j) & if\ \hat{y_{t}}(i,j,c) > \alpha_{c} \\
null & otherwise
\end{matrix}\right.
\end{split}\label{tab:eq2}
\end{equation}

where $y_{t}(i,j)$ is the pseudo label of target domain pixel $(i,j)$ and $\hat{y_{t}}$ is the output probability of the segmentation network that has $C$ channels. Applying probability constraints in self-training acts more than an inductive method, because the high-quality pseudo labels are essential in boosting the performance of instance-level and category-level adaptation.

\subsection{Triplet learning with edge pooling}
The pixel-wise feature distribution in target domain differs from source domain \cite{kouw2019review,zou2018unsupervised}, which may bring confusing predictions between classes and cause high false positive (FP) and false negative (FN) rates on the lane detection task. Moreover for lane detection models, the semantic context among the pixel of thin or dashed line instances are extremely fragile and hard to be rebuilt in the target domain. We propose to solve this in a metric learning approach with refined instance masks by edge pooling. Different from previous works on semantic segmentation UDA \cite{laradji2020m}, in which triplet loss is used to align class distributions between source and target domain, we use the triplet loss in target domain to make feature refinements. In lane segmentation UDA, we treat the four lane classes as the non-background class, which needs to be distinguished from the background class. 

A triplet \{${x}^{a}$, ${x}^{p}$, ${x}^{n}$\} is composed by an anchor, a positive example, and a negative example. In this work, an anchor or a positive example is masked features of a lane instance in non-background class, while a negative example is features of a line detected in background class. An embedding module $f$, which consists of an average pooling layer and a fully-connected layer, is used to generate l2-normalized embedding vectors \{$f({x}^{a})$, $f({x}^{p})$, $f({x}^{n})$\} from masked feature maps. Let $M, N$ be the total number of anchors and negative examples in a mini-batch, respectively. Following \cite{hermans2017defense,schroff2015facenet}, for each iteration, we utilize all possible combinations of triplets over a mini-batch to get the triplet loss:

\begin{equation}
\begin{split}
L^{t}_{tl} = \lambda_{tl} & \frac{1}{M(M-1)N}\sum_{i}^{M}\sum_{_{j}^{j \neq i}}^{M}\sum_{k}^{N} \\ & \text{max}  \{ 0, \left \| f({x_i}^{a})-f({x_j}^{p}) \right \| _{2}^{2} \\ & -\left \| f({x_i}^{a})-f({x_k}^{n}) \right \| _{2}^{2}+\beta \}
\label{tab:eq3}
\end{split}
\end{equation}

\begin{figure}[t]
  \centering
  \includegraphics[width=\linewidth]{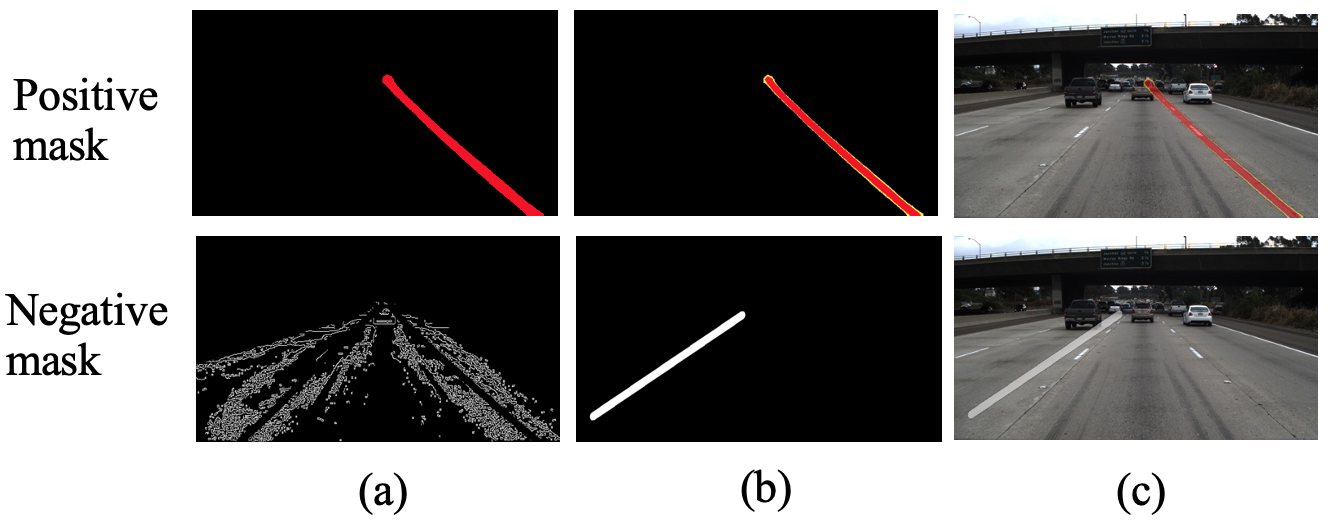}
  \caption{
  \textbf{Examples of positive instances and negative instances.} In the top row, from left to right are (a) pseudo labels, (b) expanded lane mask with edge pooling (yellow area) and (c) visualization on original image. In the bottom row, from left to right are ROI edges, negative mask and visualization.
  }
  \label{tab:fig5}
\vspace{-0.4cm}
\end{figure}

We make use of pseudo labels as well as conventional line detection methods to generate binary masks, as illustrated in Figure~\ref{tab:fig5}. In detail, we generate anchor masks with detected lanes in the pseudo labels. Then the masks are element-wise multiplied with feature maps to get anchors. Positive examples are the same as anchors. To ensure the effectiveness of triplet loss, the negative examples sampled from the background features need to be confused enough with positive examples. To achieve this, we leverage the Canny edge detector to detect edges in the raw input image and create a driving area mask by the detected lanes in the pseudo label as the region of interest  (ROI). The probabilistic hough transform (PHT) \cite{kiryati1991probabilistic} is used to extract lines from the ROI edges and obtain the negative masks.

\begin{figure}[t]
  \centering
  \includegraphics[width=0.75\linewidth]{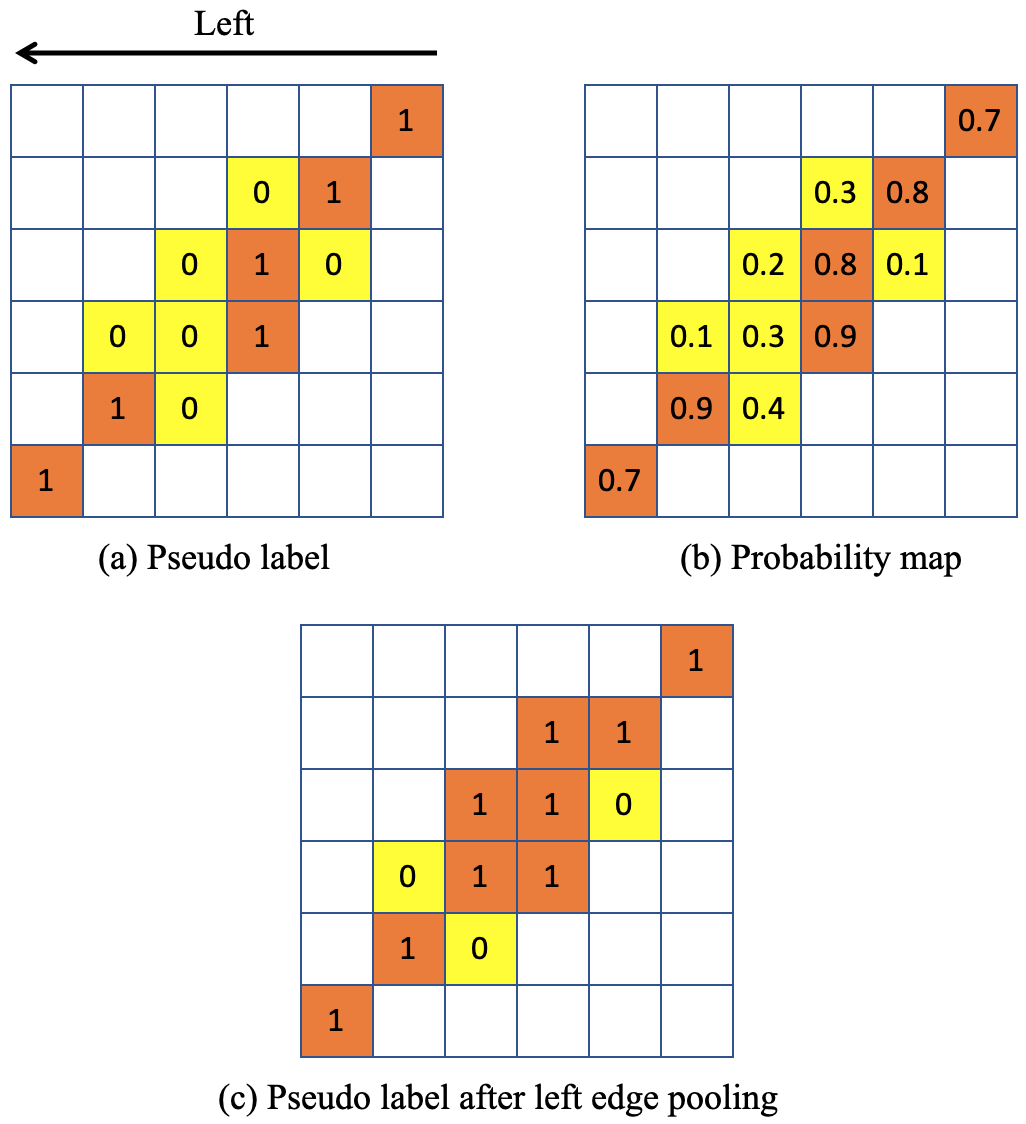}
  \caption{\textbf{An example of left edge pooling}. It scans from right to left to expand lane masks with adjacent pixels which are beyond a probability threshold (e.g. 0.2). In practice we use edge pooling in 4 directions (up, down, left, and right).}
  \label{tab:fig6}
\vspace{-0.4cm}
\end{figure}

\textbf{Edge pooling.} In previous works \cite{hermans2017defense, schroff2015facenet}, positive examples are obtained from the groundtruth of training data. In UDA, we can only generate positive examples from the pseudo labels of target domain. However, the lane masks obtained from the pseudo labels may suffer low quality because of the low confidence, which has negative effects on feature extraction and embedding generating. Inspired by corner pooling \cite{law2018cornernet}, we propose edge pooling to adaptively expand lane areas in four directions of up, down, left, and right. Edge pooling can fill the disconnected lane areas and enhance the edge of the lanes according to the probability of pixels adjacent to the pseudo labels. Thus, the lanes obtained by edge pooling are more accurate. Compared with dilation operation with fixed kernel in image morphology, edge pooling can choose the appropriate dilation rates for the near and far side of lanes so that the positive masks after edge pooling can better fit the shape of the actual lane areas. The definition of edge pooling is as follows:

\begin{equation}
t_{ij}= \left\{\begin{matrix}
1 & \text{if} \ 0 < i < H, f_{t_{(i-1)j}}>0,  p_{t_{ij}}>\gamma \\ 
0 & otherwise
\end{matrix}\right.
\label{tab:eq4}
\end{equation}

\begin{equation}
l_{ij}= \left\{\begin{matrix}
1 & \text{if} \ 0 < j < W, f_{l_{i(j-1)}}>0,  p_{l_{ij}}>\gamma \\ 
0 & otherwise
\end{matrix}\right.
\label{tab:eq5}
\end{equation}

where $f_{t}$ and $f_{l}$ are pseudo labels after one-hot preprocess, $f_{t_{ij}}$ and $f_{l_{ij}}$ are the vectors at location $(i,j)$ in $f_{t}$ and $f_{l}$ respectively. $p_{t}$ and $p_{l}$ are probablity maps, $p_{t_{ij}}$ and $p_{l_{ij}}$ are the vectors at location $(i,j)$ in $p_{t}$ and $p_{l}$ respectively. Each pseudo label in $(i,j)$ of $t_{ij}$ is 1 if the feature map in $p_{t_{ij}}$ is larger than $\gamma$ and the pseudo label in $f_{t_{(i-1)j}}$ is 1. Each pseudo label in $(i,j)$ of $l_{ij}$ is 1 if the feature map in $p_{l_{ij}}$ is larger than $\gamma$ and the pseudo label in $f_{l_{i(j-1)}}$ is 1. Edge pooling expands lane areas in four directions of up, down, left, and right. The left edge pooling is shown in Figure~\ref{tab:fig6}.

\subsection{Adaptive inter-domain embedding module}
Although the instance-wise adaptation helps alleviate feature confusion, there are still some misclassified points between lanes, and the predicted position of the lane instance will be disordered in some cases. We propose to solve this at category level with an adaptive inter-domain embedding module and a multi-label classification loss function on the existence of lane categories. The adaptive inter-domain embedding module includes channel and spatial attention sub-modules designed in Figure~\ref{tab:fig4}. Specifically, the channel-wise module reduces misclassification of points and the spatial-wise module utilizes the position prior of lanes.

Formally, we define the input feature as $x\in R^{H \times W \times C_{1}}$, where H, W and C are the height, width and channel of input feature respectively. In spatial-wise statistics $f_{spa}$, we design large-kernel convolutions to establish a close link between the feature maps and the category, making the category easier to distinguish at spatial level. Specifically, a symmetrical and separable large filter that employs a combination of $1 \times 7 + 7 \times 1$ and $7 \times 1 + 1 \times 7$ convolutions are used to reduce model parameters and obtain dense connections in a $7 \times 7$ area. After that, a spatial descriptor is adopted to represent global chancel information by Softmax. We can get the adaptive spatial-wise weights with the shape of $H \times W \times 1$ in each channel level. As for channel-wise statistics $f_{cha}$, a channel descriptor is adopted to represent global spatial information by global average pooling and global max pooling to obtain global information embedding. Then, we use Sigmoid to get the adaptive weights of channel level with the shape of $1 \times 1 \times C_{2}$ in each spatial level. The compact representation of adaptive inter-domain embedding module $f_{inter}$ is as follows:

\begin{equation}
x_{n} = f_{bn}\left ( x  \right ), n = 1, 2
\label{tab:eq6}
\end{equation}

\begin{equation}
f_{inter}\left ( x \right )= \left [ f_{cha}\left ( x_{1} \right ) \odot x_{1} \right ] \otimes \left [ f_{spa}\left ( x_{2} \right ) \odot x_{2} \right ]
\label{tab:eq7}
\end{equation}

$f_{bn}$ shares weights of three convolutions after the backbone of network with the shape of $H \times W \times C_{2}$. $\odot$ and $\otimes$ are Hadamard product for element-wise multiplication and matrix multiplication respectively. We reshape the two-dimensional feature of $f_{inter} \in R^{C_{2} \times C_{2}}$ to one-dimensional feature. Then, the fully connected (FC) layer and Sigmoid function are adopted in turn to obtain the existence of four lanes. Finally we optimize the multi-label classification loss $L_{ce}$ with self-training:

\begin{equation}
\begin{split}
z_{t} = \left\{\begin{matrix}
c & if\ f_{s}(\hat{z_{t}}(c)) > \eta \\
null & otherwise
\end{matrix}\right.
\end{split}\label{tab:eq8}
\end{equation}

\begin{equation}
L^{t}_{ce} = \lambda_{ce} \sum_{t \in T}^{ } CE(z_{t},\hat{z_{t}})
\label{tab:eq9}
\end{equation}

where $\hat{z_{t}}$ and $z_{t}$ are outputs of FC and pseudo labels of lane classes respectively. $f_{s}$ is the Sigmoid function and $CE$ represents category-wise cross-entropy loss.

To sum up, the final loss function for the unsupervised domain adaption is as follows,

\begin{equation}
\begin{split}
L=L^{s}_{seg} + L^{t}_{st} + L^{t}_{tl} + L^{t}_{ce}
\end{split}\label{tab:eq10}
\end{equation}

\section{Experiments}
In this section, we conduct extensive experiments between difficult and simple lane scenes to prove the effectiveness of our approach for cross-domain lane detection. We compare MLDA with several representative state-of-the-art methods in the research fields of UDA. ADVENT \cite{vu2019advent} combines adversarial training with entropy minimization. PyCDA \cite{lian2019constructing} is based on CDA \cite{zhang2019curriculum} and self-training. Maximum squares loss \cite{chen2019domain} balances the gradient of well-classified target samples compared with entropy minimization \cite{vu2019advent}. It should be noted that all the results are obtained without any model ensemble strategy.

\begin{table*}
  \caption{Quantitative comparison on ``TuSimple to CULane”. ``Source only” denotes directly applying the model trained on TuSimple \cite{TuSimple} to CULane \cite{pan2018spatial} without adaptation. ``Target only'' denotes supervised training on target domain. For crossroad, it only shows FP.}
  \label{tab:table1}
  \resizebox{\textwidth}{!}{
  \begin{tabular}{lcccccccccc}
    \toprule
    Experiment Setting&Normal&Crowded&Night&No line&Shadow&Arrow&Dazzle light&Curve&Crossroad&Total\\
    \midrule
Source only&50.0&25.5&19.6&18.5&16.8&36.0&25.4&34.2&7405&30.5\\
Target only&91.9&71.0&69.1&46.6&74.2&87.3&65.6&69.7&2632&73.8\\
\midrule
Advent~\cite{vu2019advent}&49.3&24.7&20.5&18.4&16.4&34.4&26.1&34.9&6527&30.4\\
PyCDA~\cite{lian2019constructing}&41.8&19.9&13.6&15.1&13.7&27.8&18.2&29.6&\textbf{4422}&25.1\\
Maximum Squares~\cite{chen2019domain}&50.5&27.2&20.8&19.0&20.4&40.1&27.4&38.8&10324&31.0\\
\textbf{Ours (MLDA)}&\textbf{61.4}&\textbf{36.3}&\textbf{27.4}&\textbf{21.3}&\textbf{23.4}&\textbf{49.1}&\textbf{30.3}&\textbf{43.4}&11386&\textbf{38.4}\\		
  \bottomrule
\end{tabular}}
\end{table*}

\subsection{Dataset}

We focus on the domain adaptation of lane detection and design the standard benchmark settings including ``TuSimple to CULane'' and ``CULane to TuSimple'' in the experiments. ``TuSimple to CULane'' means that the source domain is TuSimple \cite{TuSimple} and the target domain is CULane \cite{pan2018spatial} that is the scene from simple to difficult on domain adaptation.

\par {\textbf{CULane} \cite{pan2018spatial} is a large scale dataset for lane detection containing more than 55 hours of videos, and 133,235 frames are extracted from them. The dataset splits 88880, 9675 and 34680 frames for training set, validation set and test set, respectively. The test set is divided into the normal class and 8 challenging classes including crowded, night, no line, shadow, arrow, dazzle light, curve and crossroad. In this dataset, a total of four lane classes are set up.}

\par {\textbf{TuSimple} \cite{TuSimple} is a small scale dataset for lane detection. It mainly collects camera videos on the highway with annotated lane markings. It has approximately 7,000 one-second-long video clips of 20 frames each, and the last frame of each clip contains labeled lanes. Most frames contain four lane classes and few contain five.}

\subsection{Implementation Details}

In the implementations, we employ PyTorch deep learning framework~\cite{paszke2017automatic}. All experiments are performed on 8 NVIDIA 1080Ti GPUs. Regarding the data pre-processing, all the images are resized to 768$\times$256 and the augmentation is only 1-degree rotation. We choose ERFNet~\cite{romera2017erfnet} pre-trained on ImageNet \cite{russakovsky2015imagenet} as network backbone.

\textbf{Training.}\ The source domain model is trained with the source images by 12 epochs. We use the SGD optimizor \cite{bottou2010large} with batch size 80, learning rate 0.01, momentum of 0.9 and weight decay 1e-4. To schedule the learning rate, we follow $lr=lr\times(1-\frac{cur\_epoch}{epoch\_num})^{0.9}$. For domain adaptation training, we set initial learning rate to 0.001. For pixel-level adaptation (PL) , we choose 0.3 and 0.8 as confidence constraints for non-background and background classes. For instance-level adaptation (IL), we set $\beta$ to 1.0 and choose 0.45 as the confidence threshold $\gamma$ for four directions of edge pooling. For category-level adaptation (CL), the threshold $\eta$ for the existence of each lane class is set to 0.7. To balance the loss items in Equation~\ref{tab:eq10}, we set $\lambda_{st}$, $\lambda_{tl}$ and $\lambda_{ce}$ to 1.0, 1.0 and 0.1 respectively.

\textbf{Evaluation.}\ We directly adopt the evaluation code released alongside with CULane \cite{pan2018spatial} and TuSimple \cite{TuSimple}. The evaluation formula of the CULane is $\text{F1-measure} = \frac{2 \times \text{precision} \times \text{recall}}{\text{precision} + \text{recall}}$, where $precision=\frac{TP}{TP + FP}$, $recall=\frac{TP}{TP + FN}$. As for TuSimple, $accuracy= \frac{\sum_{clip} C_{clip}}{\sum_{clip} S_{clip}}$, $FP= \frac{F_{pred}}{N_{pred}}$ and $FN= \frac{M_{pred}}{N_{gt}}$ are used, where $C_{clip}$ and $S_{clip}$ 
represent correct points and requested points in the last frame of the clip. $F_{pred}$, $N_{pred}$,  $M_{pred}$ and $N_{gt}$ are wrong predicted lanes, predicted lanes, unannotated lanes and annotated lanes in the predictions, respectively. The segmentation masks are resized to the original image size before evaluation.

\subsection{Results on ``TuSimple to CULane''}
To verify the effectiveness of MLDA, we use TuSimple \cite{TuSimple} as the source domain and CULane \cite{pan2018spatial} as the target domain to conduct experiments, which can be considered as domain adaptation from simple scenes to difficult scenes. Note that all the methods in the Table~\ref{tab:table1} use ERFNet \cite{romera2017erfnet} as the network backbone. With the same source models and optimized training settings, we take out the best results on the target dataset for the listed methods. Table~\ref{tab:table1} summarizes the results, indicating that our method is capable of adapting simple scenes to difficult scenes and has a competitive advantage compared with other methods. Our method has a considerable improvement in the normal scene because images in TuSimple are mostly collected on highways, which is relatively similar to the images in the normal scene of CULane. The appearances of the lanes in other scenes are significantly different from TuSimple, which limits the performance of adaptation. In total, MLDA has improvements of 7.9\% and 7.4\% on F1-score compared with ``Source only'' and the second-best ``Maximum Squares''~\cite{chen2019domain}. Notice that methods such as PyCDA~\cite{lian2019constructing}, which use the same threshold for each class in generating the pseudo labels, suffer a performance drop if directly applied to this scene because the non-background class is suppressed by background class during the training progress. For all domain adaptation methods, we find that as the F1-score improves, the FP of the crossroad scene becomes higher, so that the metric on crossroad can not directly reflect the lane detection performance. The visualization results are shown in Figure~\ref{tab:fig3}.

\subsection{Results on ``CULane to TuSimple''}
Table~\ref{tab:table2} summarizes the experimental results on the domain adaptation scene of  ``CULane to TuSimple'', compared with well-performing methods \cite{vu2019advent,chen2019domain,lian2019constructing}. These results demonstrate that the MLDA approach achieves the best performance on ``CULane to TuSimple'' with the lowest FP and FN. In detail, MLDA has accuracy improvements of 28.8\% and 8.8\% compared with ``Source only'' and the second-best ``PyCDA''. Our model performs the multi-level domain adaptation method in three local to overall dimensions, pixel level, instance level, and category level, which greatly improves the accuracy of the target domain. The performances of maximum squares loss \cite{chen2019domain} and Advent \cite{vu2019advent} are relatively close, but they bring the rise of FP. Although PyCDA \cite{lian2019constructing} can improve accuracy, it also brings higher FP. Figure~\ref{tab:fig3-1} shows the visualization results.

\begin{table}
  \centering
  \caption{Quantitative comparison on ``CULane to TuSimple”. ``Source only” denotes directly applying the model trained on CULane \cite{pan2018spatial} to TuSimple \cite{TuSimple} without adaptation.}
  \label{tab:table2}
  \resizebox{0.85\linewidth}{!}{
  \begin{tabular}{lccc}
    \toprule
    Experiment Setting&FP&FN&Accuracy\\
    \midrule
    Source only &31.6&55.2&60.9\\
    Target only &19.3&4.1&95.6\\
    \midrule
    Advent~\cite{vu2019advent}&39.7&43.9&77.1\\
    PyCDA~\cite{lian2019constructing}&51.9&45.1&80.9\\
    Maximum Squares~\cite{chen2019domain}&38.2&42.8&76.0\\
    \textbf{Ours (MLDA)}&\textbf{29.5}&\textbf{18.4}&\textbf{89.7}\\
  \bottomrule
\end{tabular}}
\end{table}

\subsection{Ablation study}
In order to further analyze the effectiveness of MLDA, we use ``CULane to TuSimple'' scene for ablation research, i.e., taking CULane \cite{pan2018spatial} as the source domain and TuSimple \cite{TuSimple} as the target domain. In this case, we evaluate the effectiveness of MLDA through method stacking. As shown in Table~\ref{tab:table3}, the MLDA can significantly improve the performance in cross-domain lane detection. Pixel-level adaptation (PL) bridges the source domain and target domain to reduce the impact of confidence imbalance, making an improvement of 23.7\% on accuracy compared with source model. Instance-level adaptation (IL) makes feature refinements of the shape of lanes on target domain only so that combining PL with IL brings a considerable drop (11\%) on the FP rate and a further improvement (9.8\%) on the FN rate. Benefit from category-level adaptation (CL) which focuses on the position of lanes in both source domain and target domain, MLDA with PL, IL and CL achieves state-of-the-art with 89.7\% accuracy.

\begin{table}
  \centering
  \caption{Ablation study of MLDA on ``CULane to TuSimple”.}
  \label{tab:table3}
  \resizebox{\linewidth}{!}{
  \begin{tabular}{lcccccc}
    \toprule
    &PL&IL&CL&FP&FN&Accuracy\\
    \midrule
    Source only&&&&31.6&55.2&60.9\\
    &$\surd$&&&40.7&34.8&84.6\\
    &$\surd$&$\surd$&&29.7&25.0&86.9\\ 
    &$\surd$&$\surd$&$\surd$&\textbf{29.5}&\textbf{18.4}&\textbf{89.7}\\
  \bottomrule
\end{tabular}}
\end{table}

\section{Conclusion}
In this paper, we provide a new perspective in cross-domain lane detection by proposing the MLDA framework, which consists of three complementary levels of adaptation including pixel, instance and category. In pixel-level adaptation, self-training with confidence constraint balances non-background and background classes and contributes to recover the distribution in target domain. In instance-level adaptation, triplet learning with edge pooling carries out lane feature refinements at instance level to utilize the shape prior of lanes, which effectively rebuilds the semantic context of thin lanes. Furthermore in category-level adaptation, the adaptive inter-domain embedding module utilizes the position prior of lanes and integrates the existences of lanes with self-training. Experiments on the adaptation between CULane and TuSimple datasets show that our method can achieve state-of-the-art performance in simple-to-difficult and difficult-to-simple domain adaptation tasks. As a limitation, in the cases of complex scenes with strong occlusion or shadows, discontinuity still exists in the prediction of lanes, which will be involved in our future work.

{\small
\bibliographystyle{ieee_fullname}
\bibliography{Bibliography-File}

\begin{thebibliography}{10}\itemsep=-1pt

\bibitem{aly2008real}
Mohamed Aly.
\newblock Real time detection of lane markers in urban streets.
\newblock In {\em 2008 IEEE Intelligent Vehicles Symposium}, pages 7--12. IEEE,
  2008.

\bibitem{bottou2010large}
L{\'e}on Bottou.
\newblock Large-scale machine learning with stochastic gradient descent.
\newblock In {\em Proceedings of COMPSTAT'2010}, pages 177--186. Springer,
  2010.

\bibitem{chen2017deeplab}
Liang-Chieh Chen, George Papandreou, Iasonas Kokkinos, Kevin Murphy, and Alan~L
  Yuille.
\newblock Deeplab: Semantic image segmentation with deep convolutional nets,
  atrous convolution, and fully connected crfs.
\newblock {\em IEEE transactions on pattern analysis and machine intelligence},
  40(4):834--848, 2017.

\bibitem{chen2017rethinking}
Liang-Chieh Chen, George Papandreou, Florian Schroff, and Hartwig Adam.
\newblock Rethinking atrous convolution for semantic image segmentation.
\newblock {\em arXiv preprint arXiv:1706.05587}, 2017.

\bibitem{chen2018encoder}
Liang-Chieh Chen, Yukun Zhu, George Papandreou, Florian Schroff, and Hartwig
  Adam.
\newblock Encoder-decoder with atrous separable convolution for semantic image
  segmentation.
\newblock In {\em Proceedings of the European conference on computer vision
  (ECCV)}, pages 801--818, 2018.

\bibitem{chen2019domain}
Minghao Chen, Hongyang Xue, and Deng Cai.
\newblock Domain adaptation for semantic segmentation with maximum squares
  loss.
\newblock In {\em Proceedings of the IEEE International Conference on Computer
  Vision}, pages 2090--2099, 2019.

\bibitem{chen2017no}
Yi-Hsin Chen, Wei-Yu Chen, Yu-Ting Chen, Bo-Cheng Tsai, Yu-Chiang Frank~Wang,
  and Min Sun.
\newblock No more discrimination: Cross city adaptation of road scene
  segmenters.
\newblock In {\em Proceedings of the IEEE International Conference on Computer
  Vision}, pages 1992--2001, 2017.

\bibitem{cordts2016cityscapes}
Marius Cordts, Mohamed Omran, Sebastian Ramos, Timo Rehfeld, Markus Enzweiler,
  Rodrigo Benenson, Uwe Franke, Stefan Roth, and Bernt Schiele.
\newblock The cityscapes dataset for semantic urban scene understanding.
\newblock In {\em Proceedings of the IEEE conference on computer vision and
  pattern recognition}, pages 3213--3223, 2016.

\bibitem{everingham2015pascal}
Mark Everingham, SM~Ali Eslami, Luc Van~Gool, Christopher~KI Williams, John
  Winn, and Andrew Zisserman.
\newblock The pascal visual object classes challenge: A retrospective.
\newblock {\em International journal of computer vision}, 111(1):98--136, 2015.

\bibitem{french2017self}
Geoffrey French, Michal Mackiewicz, and Mark Fisher.
\newblock Self-ensembling for visual domain adaptation.
\newblock {\em arXiv preprint arXiv:1706.05208}, 2017.

\bibitem{ganin2014unsupervised}
Yaroslav Ganin and Victor Lempitsky.
\newblock Unsupervised domain adaptation by backpropagation.
\newblock {\em arXiv preprint arXiv:1409.7495}, 2014.

\bibitem{garnett20193d}
Noa Garnett, Rafi Cohen, Tomer Pe'er, Roee Lahav, and Dan Levi.
\newblock 3d-lanenet: end-to-end 3d multiple lane detection.
\newblock In {\em Proceedings of the IEEE International Conference on Computer
  Vision}, pages 2921--2930, 2019.

\bibitem{ghafoorian2018gan}
Mohsen Ghafoorian, Cedric Nugteren, N{\'o}ra Baka, Olaf Booij, and Michael
  Hofmann.
\newblock El-gan: Embedding loss driven generative adversarial networks for
  lane detection.
\newblock In {\em Proceedings of the European Conference on Computer Vision
  (ECCV)}, pages 0--0, 2018.

\bibitem{grandvalet2005semi}
Yves Grandvalet and Yoshua Bengio.
\newblock Semi-supervised learning by entropy minimization.
\newblock In {\em Advances in neural information processing systems}, pages
  529--536, 2005.

\bibitem{hermans2017defense}
Alexander Hermans, Lucas Beyer, and Bastian Leibe.
\newblock In defense of the triplet loss for person re-identification.
\newblock {\em arXiv preprint arXiv:1703.07737}, 2017.

\bibitem{hillel2014recent}
Aharon~Bar Hillel, Ronen Lerner, Dan Levi, and Guy Raz.
\newblock Recent progress in road and lane detection: a survey.
\newblock {\em Machine vision and applications}, 25(3):727--745, 2014.

\bibitem{hoffman2017cycada}
Judy Hoffman, Eric Tzeng, Taesung Park, Jun-Yan Zhu, Phillip Isola, Kate
  Saenko, Alexei~A Efros, and Trevor Darrell.
\newblock Cycada: Cycle-consistent adversarial domain adaptation.
\newblock {\em arXiv preprint arXiv:1711.03213}, 2017.

\bibitem{hoffman2016fcns}
Judy Hoffman, Dequan Wang, Fisher Yu, and Trevor Darrell.
\newblock Fcns in the wild: Pixel-level adversarial and constraint-based
  adaptation.
\newblock {\em arXiv preprint arXiv:1612.02649}, 2016.

\bibitem{hong2018conditional}
Weixiang Hong, Zhenzhen Wang, Ming Yang, and Junsong Yuan.
\newblock Conditional generative adversarial network for structured domain
  adaptation.
\newblock In {\em Proceedings of the IEEE Conference on Computer Vision and
  Pattern Recognition}, pages 1335--1344, 2018.

\bibitem{hou2019learning}
Yuenan Hou, Zheng Ma, Chunxiao Liu, and Chen~Change Loy.
\newblock Learning lightweight lane detection cnns by self attention
  distillation.
\newblock In {\em Proceedings of the IEEE International Conference on Computer
  Vision}, pages 1013--1021, 2019.

\bibitem{isola2017image}
Phillip Isola, Jun-Yan Zhu, Tinghui Zhou, and Alexei~A Efros.
\newblock Image-to-image translation with conditional adversarial networks.
\newblock In {\em Proceedings of the IEEE conference on computer vision and
  pattern recognition}, pages 1125--1134, 2017.

\bibitem{jung2013efficient}
Heechul Jung, Junggon Min, and Junmo Kim.
\newblock An efficient lane detection algorithm for lane departure detection.
\newblock In {\em 2013 IEEE Intelligent Vehicles Symposium (IV)}, pages
  976--981. IEEE, 2013.

\bibitem{kim2008robust}
ZuWhan Kim.
\newblock Robust lane detection and tracking in challenging scenarios.
\newblock {\em IEEE Transactions on Intelligent Transportation Systems},
  9(1):16--26, 2008.

\bibitem{kiryati1991probabilistic}
Nahum Kiryati, Yuval Eldar, and Alfred~M Bruckstein.
\newblock A probabilistic hough transform.
\newblock {\em Pattern recognition}, 24(4):303--316, 1991.

\bibitem{kouw2019review}
Wouter~Marco Kouw and Marco Loog.
\newblock A review of domain adaptation without target labels.
\newblock {\em IEEE transactions on pattern analysis and machine intelligence},
  2019.

\bibitem{laine2016temporal}
Samuli Laine and Timo Aila.
\newblock Temporal ensembling for semi-supervised learning.
\newblock {\em arXiv preprint arXiv:1610.02242}, 2016.

\bibitem{laradji2020m}
Issam~H Laradji and Reza Babanezhad.
\newblock M-adda: Unsupervised domain adaptation with deep metric learning.
\newblock In {\em Domain Adaptation for Visual Understanding}, pages 17--31.
  Springer, 2020.

\bibitem{law2018cornernet}
Hei Law and Jia Deng.
\newblock Cornernet: Detecting objects as paired keypoints.
\newblock In {\em Proceedings of the European Conference on Computer Vision
  (ECCV)}, pages 734--750, 2018.

\bibitem{lee2013pseudo}
Dong-Hyun Lee.
\newblock Pseudo-label: The simple and efficient semi-supervised learning
  method for deep neural networks.
\newblock In {\em Workshop on challenges in representation learning, ICML},
  volume~3, page~2, 2013.

\bibitem{li2019line}
Xiang Li, Jun Li, Xiaolin Hu, and Jian Yang.
\newblock Line-cnn: End-to-end traffic line detection with line proposal unit.
\newblock {\em IEEE Transactions on Intelligent Transportation Systems},
  21(1):248--258, 2019.

\bibitem{lian2019constructing}
Qing Lian, Fengmao Lv, Lixin Duan, and Boqing Gong.
\newblock Constructing self-motivated pyramid curriculums for cross-domain
  semantic segmentation: A non-adversarial approach.
\newblock In {\em Proceedings of the IEEE International Conference on Computer
  Vision}, pages 6758--6767, 2019.

\bibitem{long2015fully}
Jonathan Long, Evan Shelhamer, and Trevor Darrell.
\newblock Fully convolutional networks for semantic segmentation.
\newblock In {\em Proceedings of the IEEE conference on computer vision and
  pattern recognition}, pages 3431--3440, 2015.

\bibitem{long2016unsupervised}
Mingsheng Long, Han Zhu, Jianmin Wang, and Michael~I Jordan.
\newblock Unsupervised domain adaptation with residual transfer networks.
\newblock In {\em Advances in neural information processing systems}, pages
  136--144, 2016.

\bibitem{narote2018review}
Sandipann~P Narote, Pradnya~N Bhujbal, Abbhilasha~S Narote, and Dhiraj~M Dhane.
\newblock A review of recent advances in lane detection and departure warning
  system.
\newblock {\em Pattern Recognition}, 73:216--234, 2018.

\bibitem{neven2018towards}
Davy Neven, Bert De~Brabandere, Stamatios Georgoulis, Marc Proesmans, and Luc
  Van~Gool.
\newblock Towards end-to-end lane detection: an instance segmentation approach.
\newblock In {\em 2018 IEEE intelligent vehicles symposium (IV)}, pages
  286--291. IEEE, 2018.

\bibitem{pan2009survey}
Sinno~Jialin Pan and Qiang Yang.
\newblock A survey on transfer learning.
\newblock {\em IEEE Transactions on knowledge and data engineering},
  22(10):1345--1359, 2009.

\bibitem{pan2018spatial}
Xingang Pan, Jianping Shi, Ping Luo, Xiaogang Wang, and Xiaoou Tang.
\newblock Spatial as deep: Spatial cnn for traffic scene understanding.
\newblock In {\em Thirty-Second AAAI Conference on Artificial Intelligence},
  2018.

\bibitem{paszke2017automatic}
Adam Paszke, Sam Gross, Soumith Chintala, Gregory Chanan, Edward Yang, Zachary
  DeVito, Zeming Lin, Alban Desmaison, Luca Antiga, and Adam Lerer.
\newblock Automatic differentiation in pytorch.
\newblock 2017.

\bibitem{richter2016playing}
Stephan~R Richter, Vibhav Vineet, Stefan Roth, and Vladlen Koltun.
\newblock Playing for data: Ground truth from computer games.
\newblock In {\em European conference on computer vision}, pages 102--118.
  Springer, 2016.

\bibitem{romera2017erfnet}
Eduardo Romera, Jos{\'e}~M Alvarez, Luis~M Bergasa, and Roberto Arroyo.
\newblock Erfnet: Efficient residual factorized convnet for real-time semantic
  segmentation.
\newblock {\em IEEE Transactions on Intelligent Transportation Systems},
  19(1):263--272, 2017.

\bibitem{russakovsky2015imagenet}
Olga Russakovsky, Jia Deng, Hao Su, Jonathan Krause, Sanjeev Satheesh, Sean Ma,
  Zhiheng Huang, Andrej Karpathy, Aditya Khosla, Michael Bernstein, et~al.
\newblock Imagenet large scale visual recognition challenge.
\newblock {\em International journal of computer vision}, 115(3):211--252,
  2015.

\bibitem{sankaranarayanan2018learning}
Swami Sankaranarayanan, Yogesh Balaji, Arpit Jain, Ser Nam~Lim, and Rama
  Chellappa.
\newblock Learning from synthetic data: Addressing domain shift for semantic
  segmentation.
\newblock In {\em Proceedings of the IEEE Conference on Computer Vision and
  Pattern Recognition}, pages 3752--3761, 2018.

\bibitem{schroff2015facenet}
Florian Schroff, Dmitry Kalenichenko, and James Philbin.
\newblock Facenet: A unified embedding for face recognition and clustering.
\newblock In {\em Proceedings of the IEEE conference on computer vision and
  pattern recognition}, pages 815--823, 2015.

\bibitem{tarvainen2017mean}
Antti Tarvainen and Harri Valpola.
\newblock Mean teachers are better role models: Weight-averaged consistency
  targets improve semi-supervised deep learning results.
\newblock In {\em Advances in neural information processing systems}, pages
  1195--1204, 2017.

\bibitem{TuSimple}
TuSimple.
\newblock Tusimple benchmark, 2017.

\bibitem{tzeng2017adversarial}
Eric Tzeng, Judy Hoffman, Kate Saenko, and Trevor Darrell.
\newblock Adversarial discriminative domain adaptation.
\newblock In {\em Proceedings of the IEEE Conference on Computer Vision and
  Pattern Recognition}, pages 7167--7176, 2017.

\bibitem{vu2019advent}
Tuan-Hung Vu, Himalaya Jain, Maxime Bucher, Matthieu Cord, and Patrick
  P{\'e}rez.
\newblock Advent: Adversarial entropy minimization for domain adaptation in
  semantic segmentation.
\newblock In {\em Proceedings of the IEEE Conference on Computer Vision and
  Pattern Recognition}, pages 2517--2526, 2019.

\bibitem{wang2018deep}
Mei Wang and Weihong Deng.
\newblock Deep visual domain adaptation: A survey.
\newblock {\em Neurocomputing}, 312:135--153, 2018.

\bibitem{wu2018dcan}
Zuxuan Wu, Xintong Han, Yen-Liang Lin, Mustafa Gokhan~Uzunbas, Tom Goldstein,
  Ser Nam~Lim, and Larry~S Davis.
\newblock Dcan: Dual channel-wise alignment networks for unsupervised scene
  adaptation.
\newblock In {\em Proceedings of the European Conference on Computer Vision
  (ECCV)}, pages 518--534, 2018.

\bibitem{wu2019ace}
Zuxuan Wu, Xin Wang, Joseph~E Gonzalez, Tom Goldstein, and Larry~S Davis.
\newblock Ace: Adapting to changing environments for semantic segmentation.
\newblock In {\em Proceedings of the IEEE International Conference on Computer
  Vision}, pages 2121--2130, 2019.

\bibitem{yan2017mind}
Hongliang Yan, Yukang Ding, Peihua Li, Qilong Wang, Yong Xu, and Wangmeng Zuo.
\newblock Mind the class weight bias: Weighted maximum mean discrepancy for
  unsupervised domain adaptation.
\newblock In {\em Proceedings of the IEEE Conference on Computer Vision and
  Pattern Recognition}, pages 2272--2281, 2017.

\bibitem{zhang2019curriculum}
Yang Zhang, Philip David, Hassan Foroosh, and Boqing Gong.
\newblock A curriculum domain adaptation approach to the semantic segmentation
  of urban scenes.
\newblock {\em IEEE transactions on pattern analysis and machine intelligence},
  2019.

\bibitem{zhao2017pyramid}
Hengshuang Zhao, Jianping Shi, Xiaojuan Qi, Xiaogang Wang, and Jiaya Jia.
\newblock Pyramid scene parsing network.
\newblock In {\em Proceedings of the IEEE conference on computer vision and
  pattern recognition}, pages 2881--2890, 2017.

\bibitem{zou2018unsupervised}
Yang Zou, Zhiding Yu, BVK Vijaya~Kumar, and Jinsong Wang.
\newblock Unsupervised domain adaptation for semantic segmentation via
  class-balanced self-training.
\newblock In {\em Proceedings of the European conference on computer vision
  (ECCV)}, pages 289--305, 2018.

\end{thebibliography}
}

\end{document}